\documentclass{article}
\usepackage{ijcai26}

\usepackage{times}
\usepackage{soul}
\usepackage{url}
\usepackage[hidelinks]{hyperref}
\usepackage[utf8]{inputenc}
\usepackage[small]{caption}
\usepackage{graphicx}
\usepackage{amsmath}
\usepackage{booktabs}
\usepackage{array}
\usepackage{xcolor}
\usepackage{subcaption}
\usepackage[switch]{lineno}

\definecolor{colorIndustries}{HTML}{006400}
\definecolor{colorYears}{HTML}{00008B}
\definecolor{colorOther}{HTML}{A0522D}

\newsavebox{\mainimagebox}
\newcommand{\addlogo}[3][width=0.25\textwidth]{%
  \sbox{\mainimagebox}{\includegraphics[width=\textwidth]{#2}}%
  \begin{picture}(\wd\mainimagebox,\ht\mainimagebox)
    \put(0,0){\usebox{\mainimagebox}}%
    \put(\dimexpr\wd\mainimagebox-1pt\relax, 1pt){%
      \makebox[0pt][r]{\includegraphics[#1]{#3}}%
    }%
  \end{picture}%
}

\title{Can Open-Weight Models Compete on Financial Text Comprehension?}

\author{
Jan Sp\"orer
\affiliations
University of St.\,Gallen, Switzerland
\emails
jan.spoerer@whu.edu
}

\begin{document}

\maketitle

\begin{abstract}
Open-weight language models from Chinese AI labs caught up on benchmarks relative to proprietary frontier models in recent months. Yet their reliability on real-world financial tasks remains largely untested. We updated the Financial Touchstone benchmark, which now has 2,967 question-context-answer triplets across 495 international annual reports. We also apply a new set of models on the benchmark, expanding coverage from eleven to twenty models across ten providers, including recent open-weight models such as GLM~4.7, GLM~5, Kimi~K2.6, and DeepSeek~V3.2, as well as Alibaba's proprietary flagship Qwen3-Max. Anthropic's Claude Opus~4.6 achieves the highest accuracy (88.4\%), while Google's Gemini~2.5~Pro maintains the lowest hallucination rate (0.08\%). Notably, the open-weight Kimi~K2.6 ranks third in accuracy, and the non-reasoning models GLM~5 and Mistral~3 rank fourth and fifth, challenging the assumption that reasoning architectures or proprietary weights are a prerequisite for strong financial comprehension. Information retrieval remains the primary bottleneck, accounting for 48.9\% of all failures. We also document a new finding: geopolitical content filters in Chinese models refuse legitimate financial questions (0.08\% of attempts), sometimes without clear reason, and the refusal behavior depends on the access route as much as on the model. The complete dataset and evaluation framework are publicly available.
\end{abstract}

% information extraction, question answering, datasets, benchmarks, LLM evaluation, natural language processing, neural networks, language models, financial text, annual reports, financial reporting, corporate disclosure, needle in the haystack, context window, reasoning models, RAG

\section{Introduction}

    \subsection{The Contamination Problem in 2026}

        The pace of language model releases in 2025 and early 2026 has made benchmark contamination a pressing concern. Publicly available benchmarks are routinely absorbed into training corpora \cite{dengZhaoTangGersteinCohan_DataContamination_2024}. For follow-up studies like ours, this is problematic as a benchmark that was unseen at the time of its creation may become partially contaminated a few months later.

        Financial Touchstone was introduced in late 2025 by \cite{spoerer_2025} as a private-until-publication dataset. By keeping the source annual reports, questions, and golden answers off the public web until the benchmark is formally released, we retain reasonable confidence that frontier models evaluated in this paper have not been directly trained on our test items. In this work we exploit this property to produce an updated, comparable evaluation across twenty models from ten providers, and we additionally check for residual contamination by comparing models with different training cut-offs (see section~\ref{ft12:subsection:results-checks-for-training-data-contamination}).

    \subsection{Why Annual Reports Are a Demanding Domain}

        Annual reports are a suitable domain for three reasons. First, they are \emph{economically consequential}: markets react measurably to their release, with systematically higher volatility on earnings days \cite{BallKothari_SecurityReturnsAroundEarningsAnnouncements_1991,LandsmanMaydes_HadTheInformationContentOfQuarterlyEarningsAnnouncementsDeclinedInThePastThreeDecades_2002}, and both retail and institutional investors rely on them either directly or through equity research \cite{AsquithMikhailAu_InformationContentOfEquityAnalystReports_2005}. Analysts derive their price targets \cite{BoniniZanettiBianchiniSalvi_TargetPriceAccuracyInEquityResearch_2010,GleasonJohnsonLi_ValuationModelUseAndThePriceTargetPerformanceOfSellSideEquityAnalysts_2013} and investment recommendations \cite{Womack_DoBrokerageAnalystsRecommendationsHaveInvestmentValue_1996,barberLehavyMcnicholsTrueman_CanInvestorsProfitFromTheProphetsSecurityAnalystRecommendationsAndStockReturns_2001} from them. Other market participants, such as private investors, suppliers, competitors, and even employees of the respective companies use annual reports to find company-specific or market-specific information.

        Second, they are \emph{long and heterogeneous}. Our dataset contains reports that exceed 700 pages, mixing narrative strategy discussions, regulatory boilerplate, multi-page financial statements, and footnote tables that frequently span pages. A single report interleaves hundreds of financial metrics across sections, with information about the same quantity appearing in summary pages, the three primary statements, and management commentary.

        Third, they are \emph{open-domain}. A useful financial assistant must handle American, various European, Chinese, and Indian crore-denominated tables alongside many smaller jurisdictions. This combination of factors, with long, unstructured, numerically dense, and jurisdictionally heterogeneous text, makes annual report QA a harder test than the single-document, short-context benchmarks that dominated prior work.

    \subsection{Contributions}

        This paper adds quality checks (mostly LLM-based with a human in the loop) to the dataset presented in \cite{spoerer_2025} and also updates the list of models to reflect the state-of-the-art of closed- and open-weight language models in early 2026.

        \begin{itemize}
            \item \textbf{Model Updates for early 2026:} We added recent models such as GLM-5 and Gemini 3 Pro. In particular, we added multiple open-weight models to the comparison as those models became competitive with the flagship models of the American frontier AI labs in recent months. Thus, we contribute by benchmarking the new wave of (mostly Chinese) open-weight models.
            \item \textbf{Relevant Model Selection:} As in \cite{spoerer_2025}, the model selection targets the strongest and most widely used systems \cite{chiangZhengShenAngelopoulos_LLMArena_ChatBotArena_2024}, keeping the study relevant to the majority of LLM users and researchers.
            \item \textbf{Comprehensive Analysis:} We make detailed analyses of the errors that models still make on real-world financial document comprehension tasks. In particular, we also dive into refusals.
            \item \textbf{Open Data and Reproducibility:} We release all data on our website: \href{https://financial-touchstone.datascience-nlp.ai/}{financial-touchstone.datascience-nlp.ai/}
        \end{itemize}

\section{Related Work}

This paper extends the Financial Touchstone benchmark introduced in \cite{spoerer_2025}, which evaluated eleven frontier models on annual report comprehension. We expand the model set to twenty, with a particular focus on open-weight models from Chinese AI labs. In addition, we expand the dataset by 89 questions and 15 reports (from 2,878 questions over 480 reports in v1.1 to 2,967 questions over 495 reports in v1.2). As part of the v1.2 quality check, 67 were flagged for incomplete answers and 15 for hallucinations in the original golden answer, and the corresponding entries were revised. We briefly survey the three lines of work most relevant to this update: financial QA benchmarks, the rise of open-weight models in finance, and content filtering in language models.

\paragraph{Financial QA Benchmarks.}
Financial text analysis progressed from sentiment lexicons \cite{loughranMcDonald_WhenIsALiabilityNotALiability_TextualAnalysisDictionariesAnd10Ks_2011} and phrase-level classification \cite{MaloSinhaKorhonenWallenius_FinancialPhraseBank_GoodDebtOrBadDebtDetectingSemanticOrientationsinEconomicTexts_2014} to structured QA benchmarks requiring numerical reasoning. FinQA \cite{ChenChenSmileyShahBorovaLangdonMoussaBeaneHuangRoutledgeWang_FinQaADatasetOfNumericalReasoningOverFinancialData_2021} and TAT-QA \cite{zhuLeiHuangWangZhangLvFengChua_TatQa_AQuestionAnsweringBenchmarkOnAHybridOfTabularAndTextualContentInFinance_2021} test multi-step reasoning over semi-structured tables, but their contexts are limited to a few thousand tokens. FinanceBench \cite{islamKannappanKielaQianScherrerVidgen_FinanceBench_2023} moved closer to real-world conditions by posing questions against full annual reports, but its best model (GPT-4) achieved only 19\% accuracy, and its evaluation predates reasoning-capable models entirely. Our benchmark differs from all of these in scale (495 reports, 2,967 questions), international coverage (22 countries), and the inclusion of twenty models spanning the 2025--2026 state-of-the-art.

\paragraph{Open-Weight Models and Chinese AI in Finance.}
The rapid improvement of open-weight models is a defining trend of 2025--2026. DeepSeek R1 \cite{deepseekai_DeepSeek_R1_IncentivizingReasoningCapability2025deepseekr1incentivizingreasoningcapabilityInLLMsViaReinforcementLearning_2025} demonstrated that novel, easier-to-scale approaches for reinforcement learning can produce competitive reasoning capabilities. GLM-5 \cite{glmTeam_GLM5_FromVibeCodingToAgenticEngineering_2026} and the Qwen3 model family \cite{yangQwenTeam_Qwen3TechnicalReport_2025} further narrowed the gap. However, systematic evaluations of these models on financial tasks remain scarce. Prior work on LLMs in finance has focused almost exclusively on American models \cite{kimMuhnNikolaev_FinancialStatementAnalysisWithLargeLanguageModels_2024,papasotiriouSoodReynoldsBalch_AiInInvestmentAnalysisLlmsForEquityStockRatings_2024}. To our knowledge, our study is the first to benchmark Chinese open-weight models on an international financial dataset, revealing competitive accuracy (Moonshot~AI's Kimi~K2.6 ranks third overall).

\paragraph{Retrieval-Augmented Generation and Content Filtering.}
RAG \cite{LewisPerezPiktusPetroniGoyalKuttlerLewisYihRocktaschelRiedelKiela_RetrievalAugmentedGenerationForKnowledgeIntensiveNlpTasks_2020} remains essential for long-document QA despite some context windows now exceeding one million tokens \cite{googleDeepMind_Gemini25Pro_2025}, because the needle-in-a-haystack problem persists \cite{nelsonKolliasDasChadhuryDan_needlehaystackmemorybased_2024}. A separate and underexplored issue is that content safety filters in Chinese-provider models can block legitimate financial queries. We document this phenomenon empirically in this paper. To our knowledge, no prior benchmark has systematically measured the impact of content filtering on financial AI performance.

\section{Data}\label{ft12:sec:data}

    \subsection{Annual Report Collection}

        The dataset has 495 financial annual reports from public corporations worldwide: Australia, Austria, Belgium, Canada, the Cayman Islands, China, France, Germany, Hong Kong, India, Indonesia, Ireland, Japan, Malaysia, the Netherlands, Portugal, Singapore, South Korea, Spain, Switzerland, Taiwan, Thailand, the United Kingdom, and the United States. The resulting coverage spans heterogeneous disclosure regimes and accounting standards.

        Where the source material allowed, the selection was balanced across company size, country, industry sector, and reporting year (predominantly 2021--2023).

        % Data, task, and RAG descriptions in this section are adapted from the other paper.

    \subsection{Question Development}

        Our six question types are based on prior research identifying the most frequently asked questions by professional equity analysts \cite{popSpoerer_ffaq_identificationOfTheFinancialFrequentlyAskedQuestionsInFinancialReports_2025}. The prompts were originally introduced in \cite{spoerer_2025}; in v1.2 we revised all six to reduce ambiguity. The most substantive change is to \texttt{key\_financials}, where the open-ended ``key financial figures and metrics'' wording was replaced with an explicit enumeration of the expected metric families, so that both human annotators and models share a common understanding of what counts as ``key.'' The other revisions narrow open-ended phrasings (``mentioned in this text'', ``about the company's cash flow'') to specific quantities (total revenue, year-over-year growth rate, the three standard cash-flow components).

        \begin{enumerate}
        \item \textbf{key\_financials}: \emph{What are the key financial metrics (EBITDA, net income, EPS, margins, operating profit, cash flow)?}
        \item \textbf{cash\_flow}: \emph{What are the operating, investing, and financing cash flows?}
        \item \textbf{revenue}: \emph{What is the total revenue?}
        \item \textbf{revenue\_growth}: \emph{What is the year-over-year revenue growth rate?}
        \item \textbf{segments}: \emph{What are the business segments and their details?}
        \item \textbf{company\_type}: \emph{What is the legal form of incorporation (Inc., LLC, AG, etc.)?}
        \end{enumerate}

        Each prompt is prefixed with the company name and report year at inference time (e.g., ``For Nestle (2021): \ldots'') so that the retriever and the model know which document and which fiscal year is being queried.

    \subsection{Current Dataset Statistics}

        The annual report dataset contains:
        \begin{itemize}
            \small
            \item 495 annual reports.
            \item Coverage of all industry classes from the Global Industry Classification Standard (GICS).
            \item 2,967 manually annotated question-context-answer triplets.
            \item Six questions per report.
            \item About 83 million tokens of extracted financial text.
            \item Temporal span: Reporting years centered on 2021--2023, complemented by a small number of earlier and later reports.
            \item Listings from 20 stock exchanges in 22 countries, which together account for most of the world's market capitalization: Shanghai Stock Exchange (CN), Tokyo Stock Exchange (JP), National Stock Exchange (IN), Bombay Stock Exchange (IN), Hong Kong Stock Exchange (HK), Korea Exchange (KR), Taiwan Stock Exchange (TW), London Stock Exchange (GB), Xetra (DE), Toronto Stock Exchange (CA), Australian Securities Exchange (AU), Singapore Exchange (SG, with one company headquartered in the KY), Bursa Malaysia (MY), Thailand Stock Exchange (TH), Indonesia Stock Exchange (ID), NASDAQ (US), NYSE Euronext (US), Euronext (NL, FR, BE, PT, IE), the various BME exchanges (ES), and SIX (CH).
        \end{itemize}

    \subsection{Inter-Annotator Agreement, Disagreement Triage, and Analysis of Human Error Sources}

        Inspired by \cite{spoerer_2025,spoererGausHandschuh_GraphRAG_FinancialTouchstone2_2025}, we checked manually labeled data that we added to the dataset, and expanded the procedure with a model-in-the-loop step.

        First, we have a dual-annotation subsample: 46 questions were independently answered by two annotators from the same source document, and disagreements were adjudicated by reading the relevant sections of the report in detail. 
        
        Second, a model-in-the-loop sanity check: before launching the full twenty-model run, we generated answers with a subset of the models from all questions and manually reviewed every case in which the judge marked a model answer as incorrect. Whenever the review revealed that the human annotation was actually at fault, we traced the discrepancy back to the golden answer and revised it. This procedure resulted in 67 records being revised for incompleteness and 15 for containing claims not supported by the source text.

        Taken together, the dual-annotation analysis is consistent with the human baseline of 82.8\% accuracy and 2.8\% hallucination reported in \cite{spoerer_2025}, which we continue to use as the reference, and we retain the corrected, adjudicated answers as the golden reference. All model accuracies and hallucination rates reported in this paper are measured against this corrected golden set.

        The inter-annotator error analysis reveals recurring failure modes that are worth naming explicitly, because they recur in both humans and language models (see section~\ref{ft12:subsection:chinese-model-refusals} onward). Segment questions are the hardest, because companies routinely segment their business along several dimensions (such as product category, geography, and customer type) and an exhaustive answer requires naming all of them. It is easy for an annotator (and for a model) to confuse segment-level financial figures with whole-company financials, particularly when a table sits in a subsection about a single division. Cash-flow and key-financial questions fail for a different reason: these metrics are often split across multiple pages of the report, so a keyword-based PDF search can surface one half of the relevant table while missing a crucial header or a prior-page footnote.

        A specific source of difficulty for the \texttt{key\_financials} question is that there is no universally accepted definition of what counts as ``key'' in a given company's report. Many companies include a one-page financial summary titled ``Key Financials'', ``Our Performance'', ``at a Glance'', or similar. This page contains bespoke ratios such as adjusted EBITDA, organic growth, or net interest margin. Annotators who skipped this summary page and jumped to the balance sheet, income statement, and cash flow statement typically missed the metrics that the report's authors considered most important. As part of the v1.2 quality check, 40 \texttt{key\_financials} questions were re-reviewed specifically to reduce this ambiguity.

        Finally, we want to flag that our human baseline reflects a best-case scenario, not typical professional practice. Our annotators worked full-time for several weeks on the labeling task, with no time pressure and an explicit instruction to double-check every answer against the source. Real-world analysts operate under tighter deadlines, so the 82.8\% accuracy and 2.8\% hallucination rate we use as a reference should be read as an upper bound on human reliability in comparable industry workflows.

\section{Methodology}

    \subsection{Task Definition}

        \begin{figure}%[t]
            \begin{center}
                \caption{\textbf{Model Performance With 95\% Confidence Bars, Excluding Retriever Errors, Ranked by Hallucination Rate.}}
                \label{ft12:fig:performance_barchart_sufficient_retrieval.png}
                % \Description{Claude Opus 4.6 has the highest accuracy, followed by Claude Sonnet 4.6 and GLM 5. Gemini 2.5 Pro has the lowest hallucination rate. The dashed lines show the human baselines.}
                
                \includegraphics[width=8.3cm]{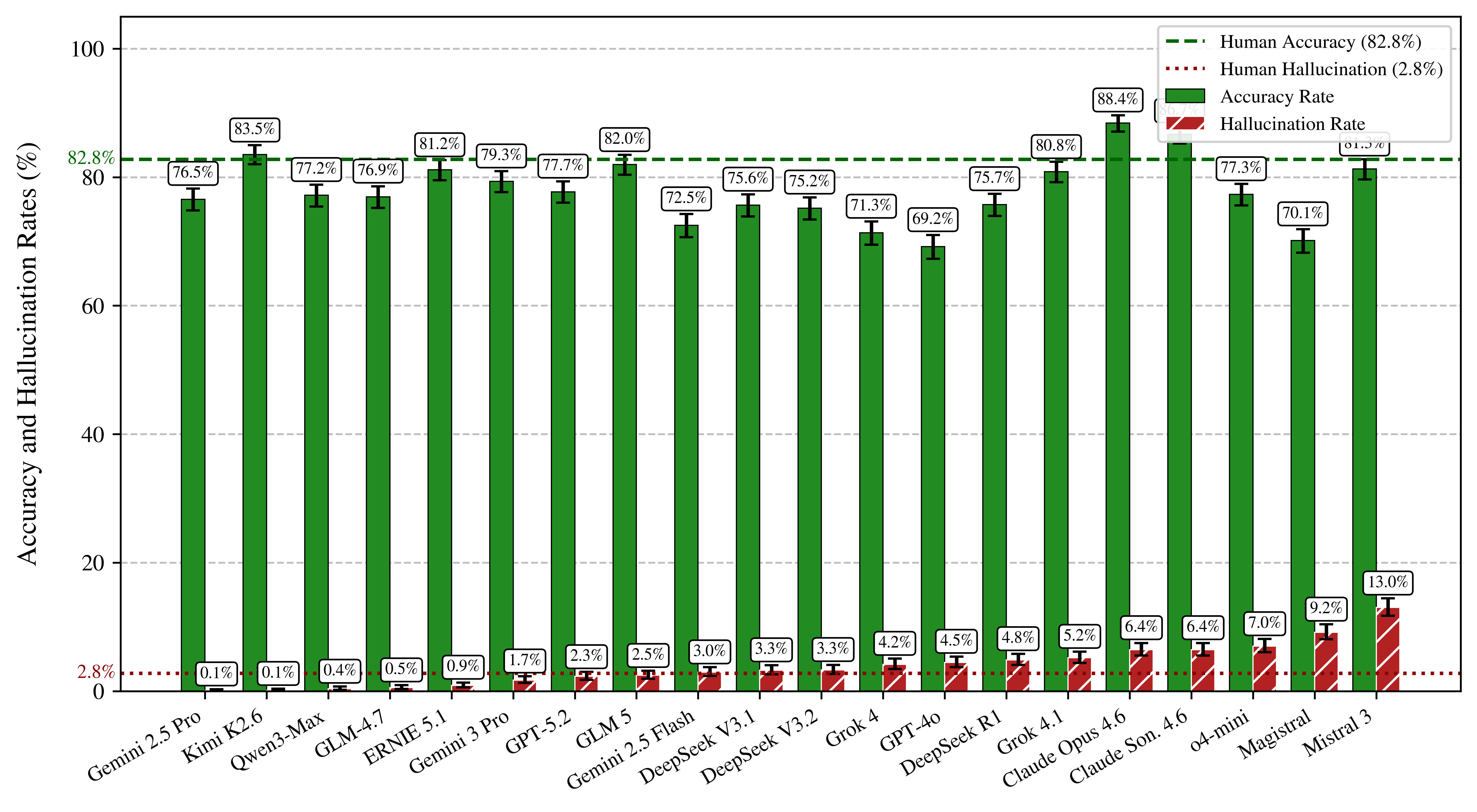}

            \end{center}
        \end{figure}

        Each benchmark item consists of a natural-language question, a full annual report as the knowledge base, and a human-curated golden answer. A model's job is to produce an answer that matches the golden answer in substance, based solely on what can be retrieved from the report. We deliberately do not provide hand-picked context: the retrieval step is part of the task, and retrieval failures are treated as failures of the end-to-end system rather than excused as an upstream issue.

        Grading is performed by an LLM judge (Claude Opus~4.6; the four models evaluated last were graded by Claude Opus~4.8 with identical instructions, see Appendix~\ref{ft12:app:judge}) that compares the model's answer against the golden answer. In v1.2, we substantially tightened the judge's instructions to reduce evaluator variance compared to \cite{spoerer_2025}. Concretely, the judge now receives (i) per-question-type correctness thresholds. For example, \texttt{key\_financials} requires the model to surface at least 60\% of the metrics in the golden answer \emph{and} all ``headline'' metrics (revenue, net income, EPS when applicable), while \texttt{segments} requires \emph{all} major segments to be listed; (ii) an explicit separation between \emph{incorrect} and \emph{hallucinated} answers, so that a misread table (numbers present in the context but assigned to the wrong label or year) counts as an error but not as a hallucination; and (iii) tolerances for rounding, unit differences, and minor naming variations. These clarifications reduce borderline judgments and make the correctness and hallucination metrics more reproducible.

    \subsection{Overview of Evaluated Closed and Open Models}\label{ft12:subsec:overview-of-evaluated-closed-and-open-models}

        We evaluate twenty models from ten open- and closed-weight providers. The following sixteen are reasoning models:
        \begin{itemize}
                \item Google Gemini 2.5 Pro, Gemini 2.5 Flash \cite{googleDeepMind_Gemini25Pro_2025}, Gemini 3 Pro \cite{googleDeepMind_Gemini3ProModelCard_2025}
                \item OpenAI o4-mini \cite{openai_o3_and_o4mini_2025}, GPT-5.2 \cite{openai_GPT5SystemCard_2025}
                \item Anthropic Claude Opus 4.6, Claude Sonnet 4.6 \cite{anthropic_claude4_2025}
                \item DeepSeek R1 \cite{deepseekai_DeepSeek_R1_IncentivizingReasoningCapability2025deepseekr1incentivizingreasoningcapabilityInLLMsViaReinforcementLearning_2025}
                \item xAI Grok 4 \cite{xai_grok4_2025}, Grok 4.1 \cite{xai_Grok41ModelCard_2025}
                \item Zhipu AI GLM-4.7 \cite{zaiOrg_GLM47_ModelCard_2025}
                \item Mistral AI Magistral \cite{rastogiMistralAI_Magistral_2025}
                \item Alibaba Qwen3-Max \cite{yangQwenTeam_Qwen3TechnicalReport_2025}
                \item Moonshot AI Kimi~K2.6 \cite{moonshotAI_KimiK26_ModelCard_2026,kimiTeam_KimiK25_VisualAgenticIntelligence_2026}
                \item Baidu ERNIE 5.1 \cite{wangErnieTeam_Ernie50TechnicalReport_2026}
                \item Zhipu AI GLM-5 (run in reasoning mode) \cite{glmTeam_GLM5_FromVibeCodingToAgenticEngineering_2026}
            \end{itemize}
        We included the following four non-reasoning models:
        \begin{itemize}
                \item OpenAI GPT-4o \cite{openai_gpt4o_2024}
                \item DeepSeek V3.1 \cite{deepseek_v3_technical_report_2025}, V3.2 \cite{deepseekAI_DeepSeekV32_PushingTheFrontierOfOpenLLMs_2025}
                \item Mistral AI Mistral 3 \cite{mistralAI_Mistral3_2025}
        \end{itemize}
        These models were chosen because they include the best-rated language model at the time according to LLM Arena \cite{chiangZhengShenAngelopoulos_LLMArena_ChatBotArena_2024}. In particular, we added multiple models from Chinese providers (DeepSeek, Zhipu AI, Alibaba, Moonshot AI, Baidu), most of them open-weight, that have become competitive with the flagship models of the American frontier AI labs. Finance-specific models \cite{WuIsroyLuDabravolskiDredzeGehrmannKambadurRosenbergMann_BloombergGPT_ALargeLanguageModelForFinance_2023} were excluded, as they no longer match the performance of large general-purpose models.

        Meta's Llama~3 \cite{touvronLavrilIzacardMartinetLachausLacroixRoziereGoyalHambroAzharRodriguezJoulinGraveLample_Llama_OpenAndEfficientFoundationLanguageModels_2023,metaAi_llama3_dot_1__2024} and Llama~4 \cite{MetaAI_TheLlama4HerdTheBeginningOfANewEraOfNativelyMultimodalAIInnovation_2025} were likewise omitted, as they trail the open- and closed-weight models of the other laboratories.

        \begin{figure}%[t]
            \begin{center}
            
                \caption{\textbf{Model Accuracy With 95\% Confidence Bars, USA vs. Rest of World.} Models ordered by hallucination rate (as in figure~\ref{ft12:fig:performance_barchart_sufficient_retrieval.png}), which is not shown here.}
            
                % \Description{Model Accuracy With 95\% Confidence Bars, USA vs. Rest of World, Ranked by Hallucination Rate (Not Displayed Here): All models work better with American 10-Ks than with other annual reports. y-axis starts at 25\%.}
                
                \includegraphics[width=8.3cm]{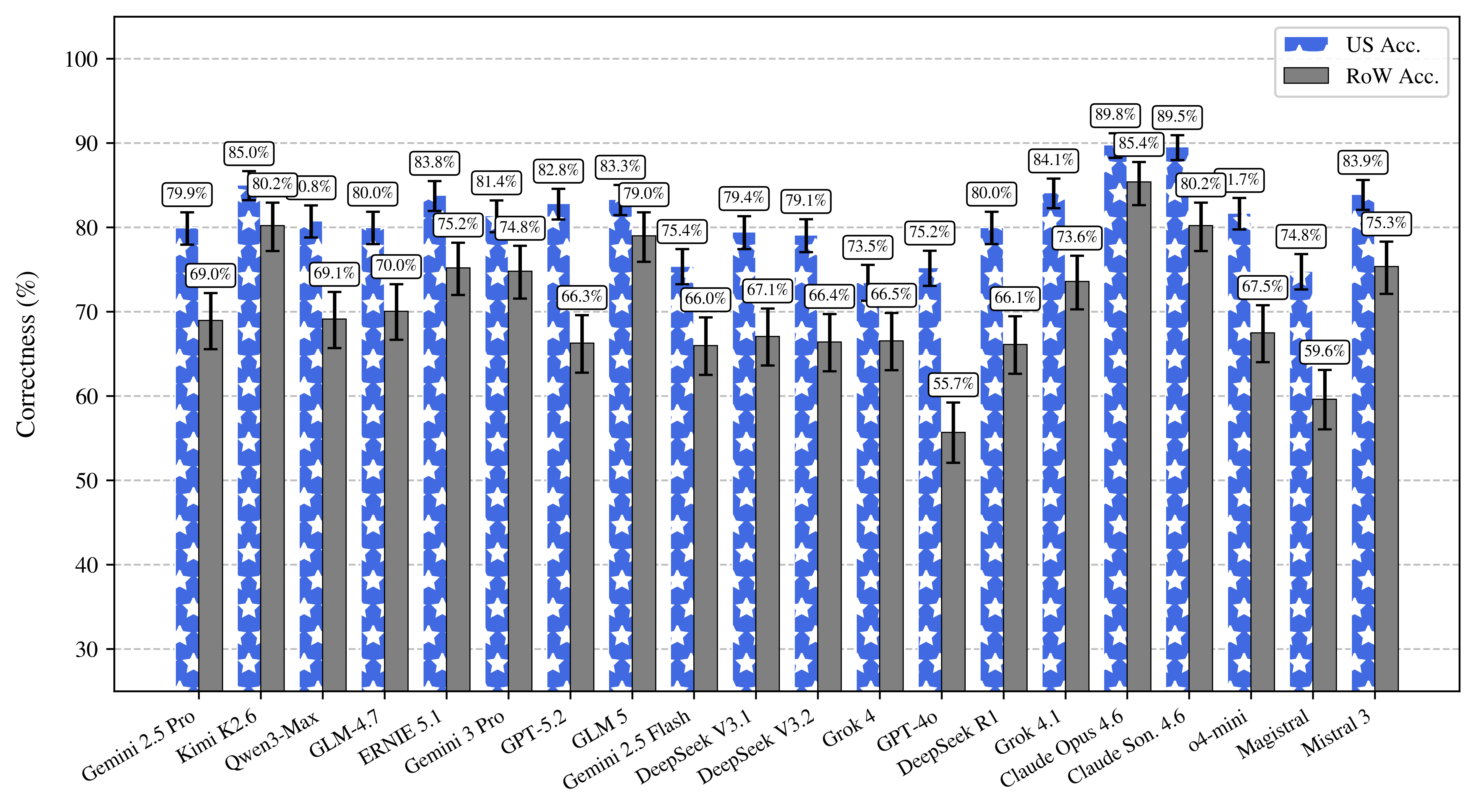}
                \label{ft12:fig:usa_vs_row_geo_performance_barchart_sufficient_retrieval.png}
            
            \end{center}
            
        \end{figure}

    \subsection{RAG Pipeline and Fairness Controls}

        Our retrieval stack is intentionally simple, both to keep the pipeline reproducible and to avoid coupling the benchmark to any provider's proprietary retrieval stack. Each report is chunked into 1000-token windows with a 200-token overlap, using OpenAI's \texttt{cl100k\_base} tokenizer. Chunks are embedded with \texttt{text-embedding-3-small}, indexed in FAISS, and queried at inference time with the same question text the model receives. We take the top-5 chunks as retrieved context and pass them, concatenated, into every model's prompt.

        We use the same retrieved context for every model. Thus, any variation in accuracy across models can be attributed to the model directly. The retriever is not a confounding variable as it is a fixed upstream condition.

        We explored expanding this pipeline (top-10 retrieval, model-specific re-ranking, per-chunk prompting) and observed no meaningful gain relative to the computational cost of running twenty models on every configuration. Top-5 remains the default. Where the retriever fails to surface the required information, we flag the item as a retrieval failure and, in downstream analyses, exclude it so that model comprehension and retrieval quality can be assessed separately (as in figures~\ref{ft12:fig:performance_barchart_sufficient_retrieval.png} and \ref{ft12:fig:usa_vs_row_geo_performance_barchart_sufficient_retrieval.png}). We revisit the retrieval bottleneck in section~\ref{ft12:subsection:results-checks-for-training-data-contamination} and table~\ref{ft12:tab:failure_taxonomy}. We acknowledge that a stronger retrieval stack (GraphRAG \cite{hanHaoyuShomerWangLeiGuoHuaLongLiuTang_RagVsGraphragASystematicEvaluationAndKeyInsights_2025}, hybrid sparse/dense retrieval, or large-context re-reading) would lift overall accuracy.

\section{Results}

        \begin{figure}%[t]
            \centering
            \caption{\textbf{Multidimensional Performance Comparison of All Models.}}

            % \Description{Each spider chart shows accuracy for industries, years, regions, and overall groundedness (i.e., inverse hallucination rate). Each circle layer is a 25\% step.}
            
            \label{ft12:fig:spider_comparison}

            % LEGACY-SLUG NOTE (2026-07-04): four subfigures below keep OLD
            % filename/label slugs from before the 2026-06-29 genuine-model
            % re-run — kimi-k2.5 (= Kimi~K2.6), glm-4.7-thinking (= GLM-4.7),
            % qwen-3 (= Qwen3-Max), ernie-5.0 (= ERNIE 5.1). The PNG contents
            % and \captions use the correct new names; only the file paths and
            % \labels are legacy, kept intentionally so nothing churns. Do NOT
            % "fix". Full map: mcp_financial_touchstone/MODEL_NAME_PROVENANCE.md
            % ("Legacy slug map") and the pipeline scripts' MODELS lists.

            % --- Row 1: Google ---
            \begin{subfigure}[b]{0.11\textwidth}
                \centering
                \addlogo{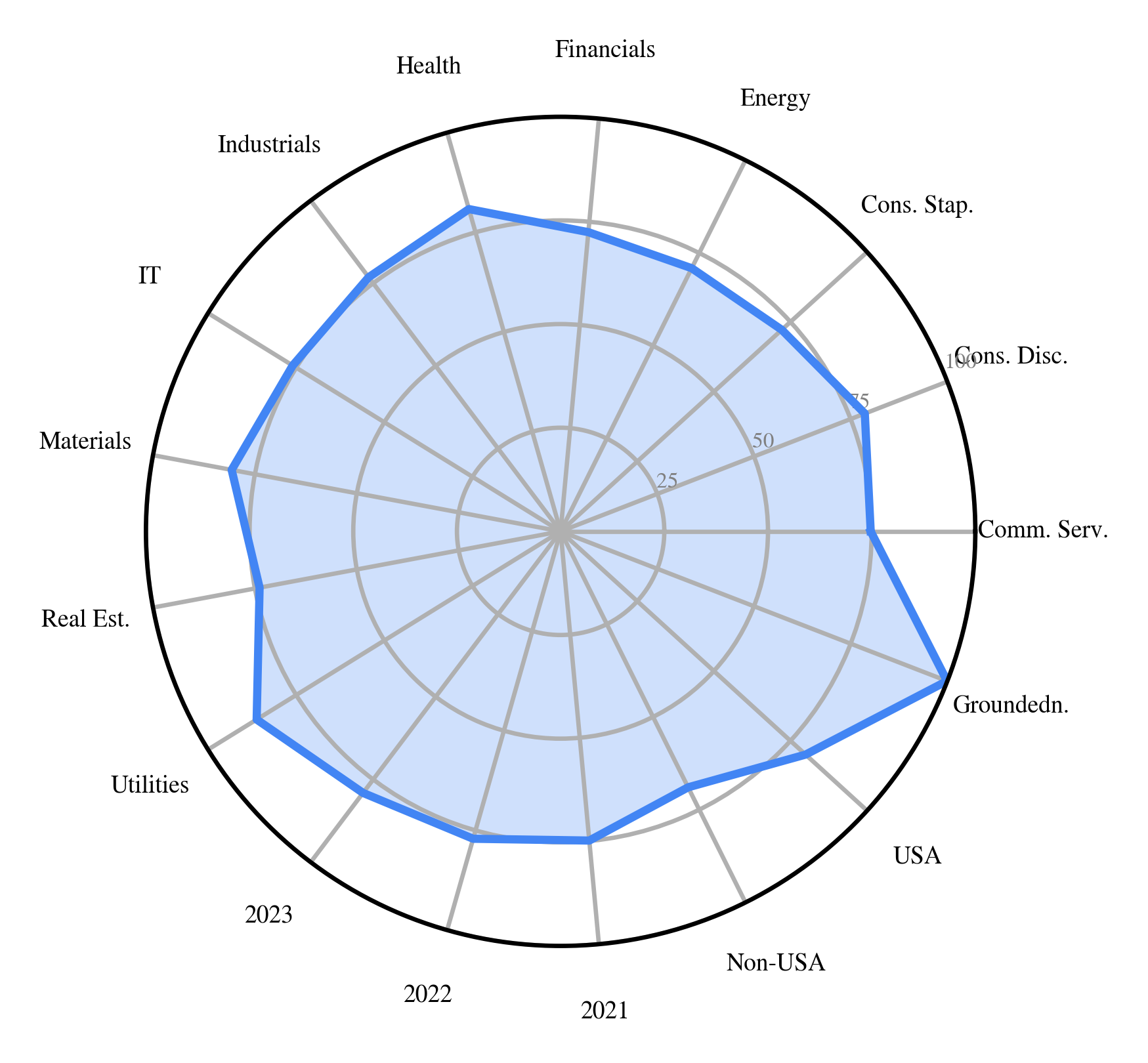}{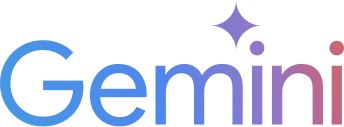}
                \caption{G.\ 2.5 Pro}
                \vspace{-0.1cm}
                \label{ft12:fig:spider_chart_gemini-2.5-pro}
            \end{subfigure}
            \begin{subfigure}[b]{0.11\textwidth}
                \centering
                \addlogo{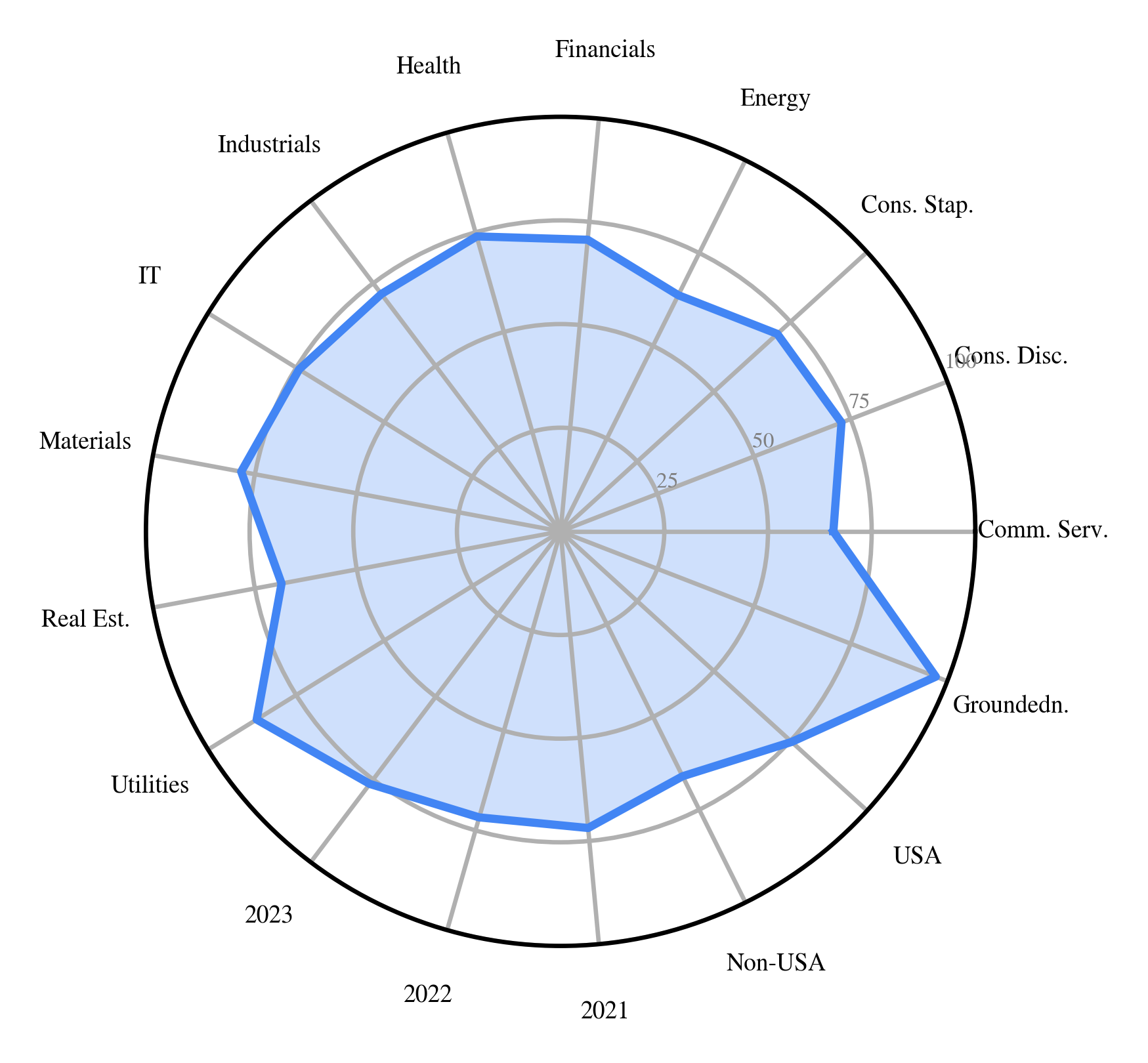}{images_final/logo_google_gemini.png}
                \caption{" 2.5 Fl.\ }
                \vspace{-0.1cm}
                \label{ft12:fig:spider_chart_gemini-2.5-flash}
            \end{subfigure}
            \begin{subfigure}[b]{0.11\textwidth}
                \centering
                \addlogo{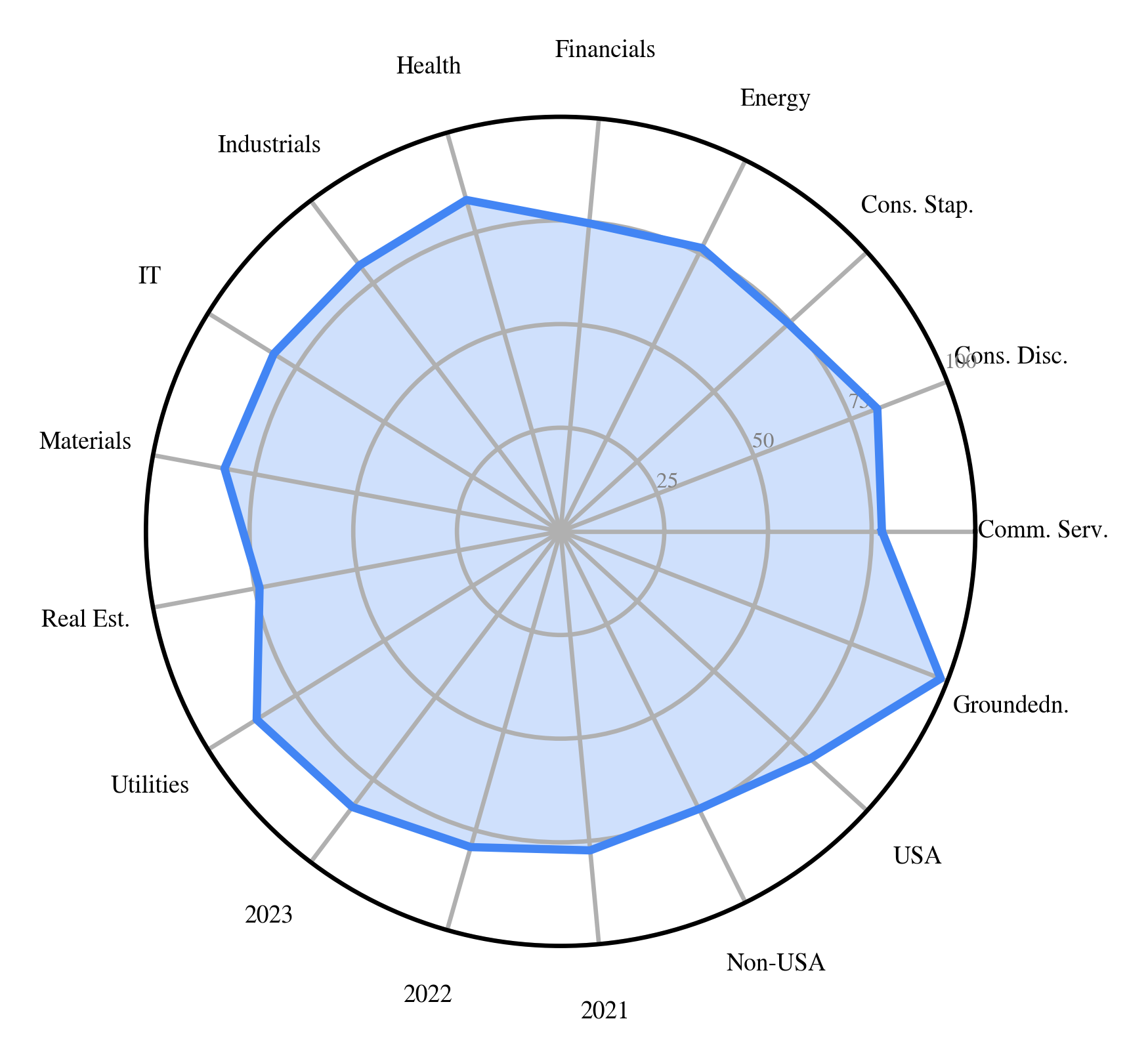}{images_final/logo_google_gemini.png}
                \caption{G.\ 3 Pro}
                \vspace{-0.1cm}
                \label{ft12:fig:spider_chart_gemini-3-pro}
            \end{subfigure}
            \begin{subfigure}[b]{0.11\textwidth}
                \centering
                \addlogo[width=0.3\textwidth]{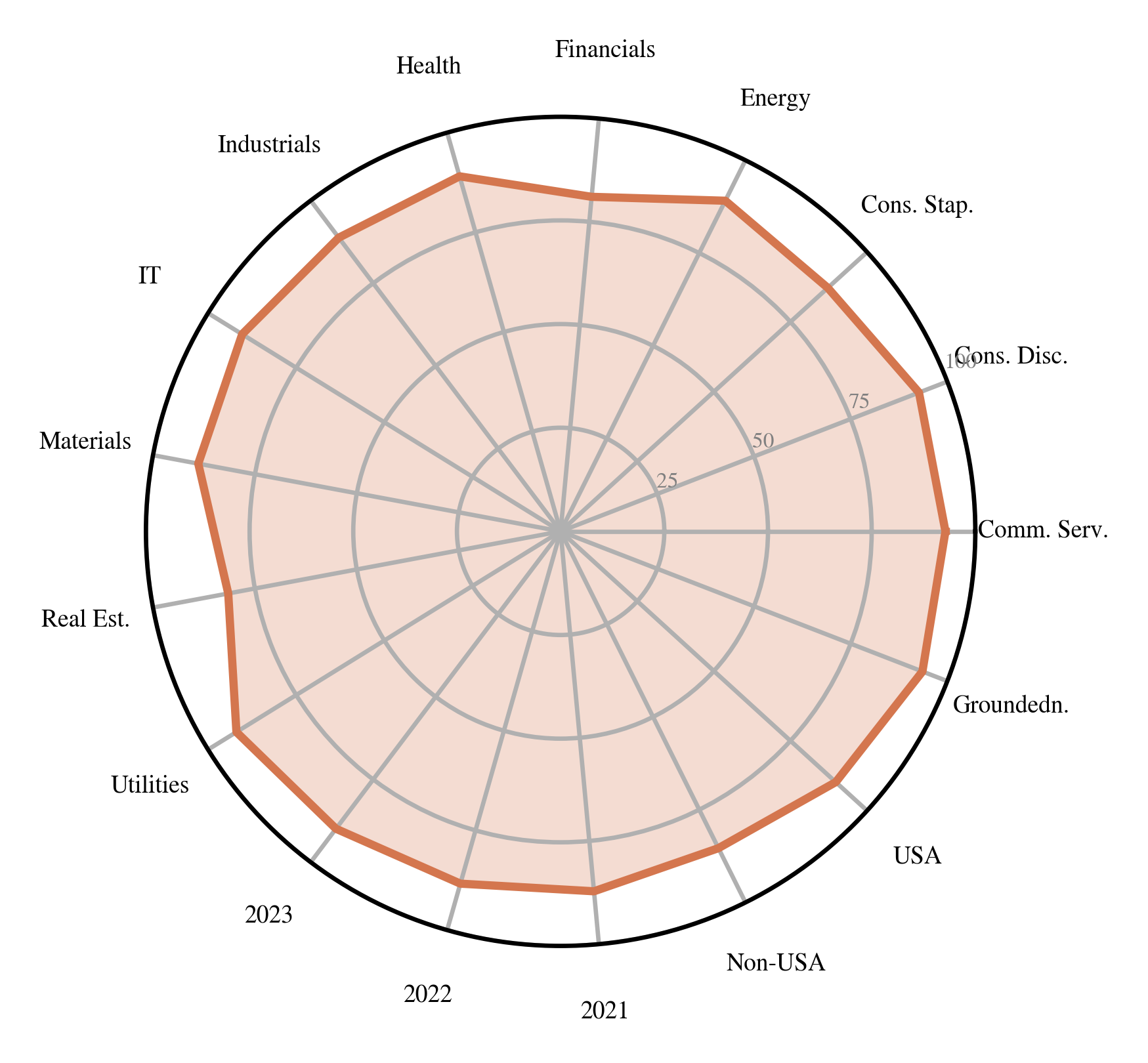}{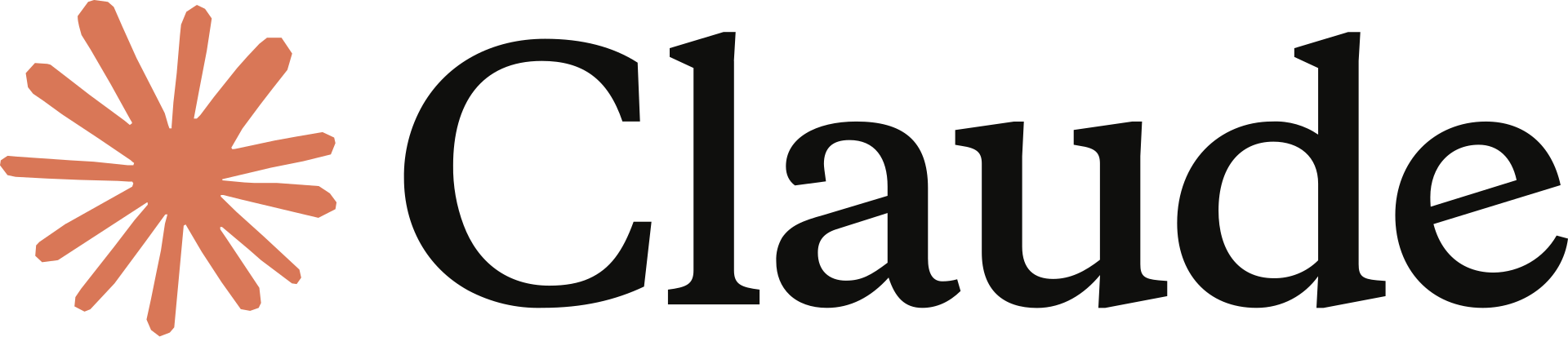}
                \caption{C.\ O.\ 4.6}
                \vspace{-0.1cm}
                \label{ft12:fig:spider_chart_claude-opus-4.6}
            \end{subfigure}

            \vspace{0.5em}

            %  Row 2: Anthropic + OpenAI 
            \begin{subfigure}[b]{0.11\textwidth}
                \centering
                \addlogo[width=0.3\textwidth]{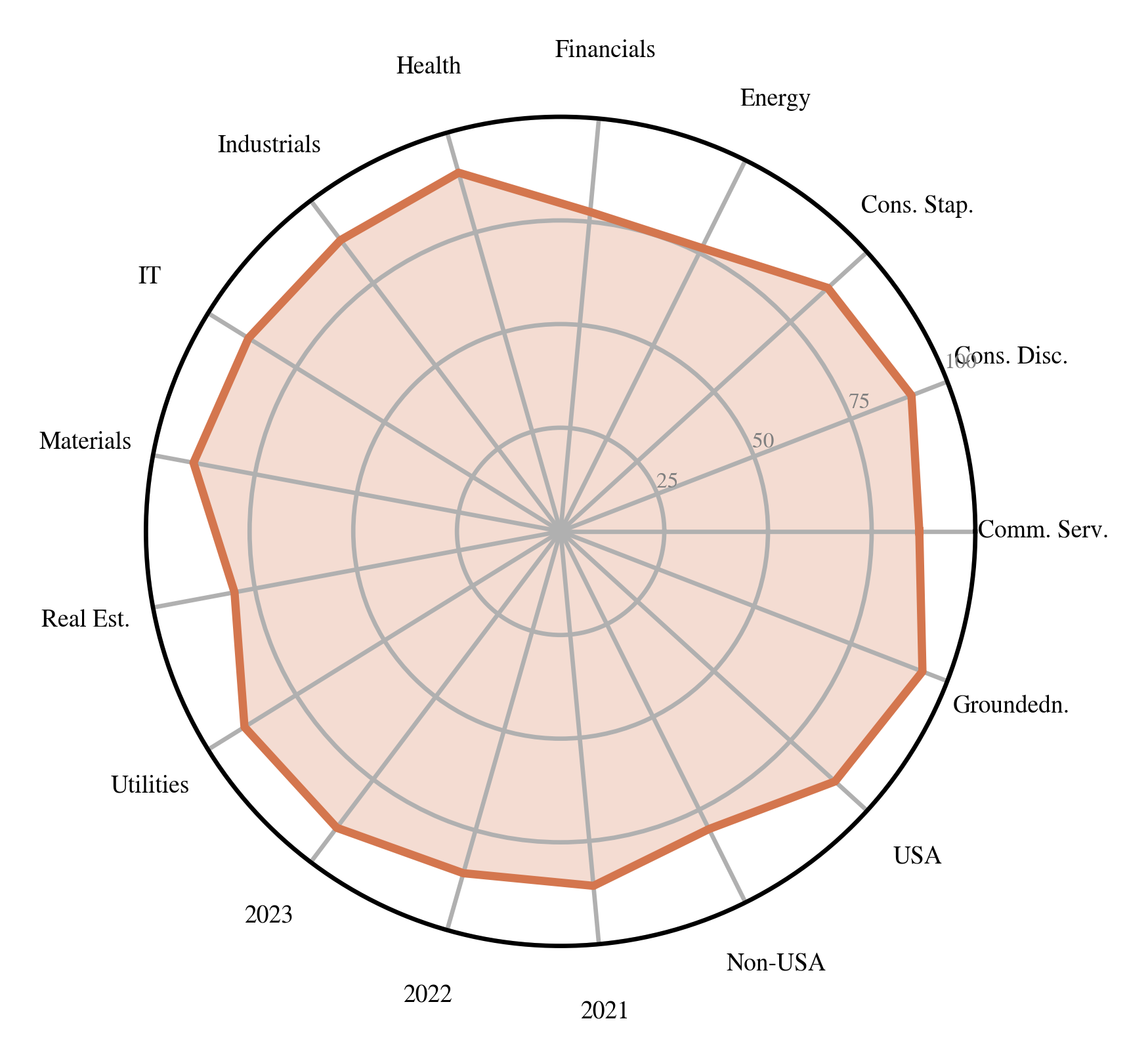}{images_final/logo_claude.png}
                \caption{" S.\ 4.6}
                \vspace{-0.1cm}
                \label{ft12:fig:spider_chart_claude-sonnet-4.6}
            \end{subfigure}
            \begin{subfigure}[b]{0.11\textwidth}
                \centering
                \addlogo{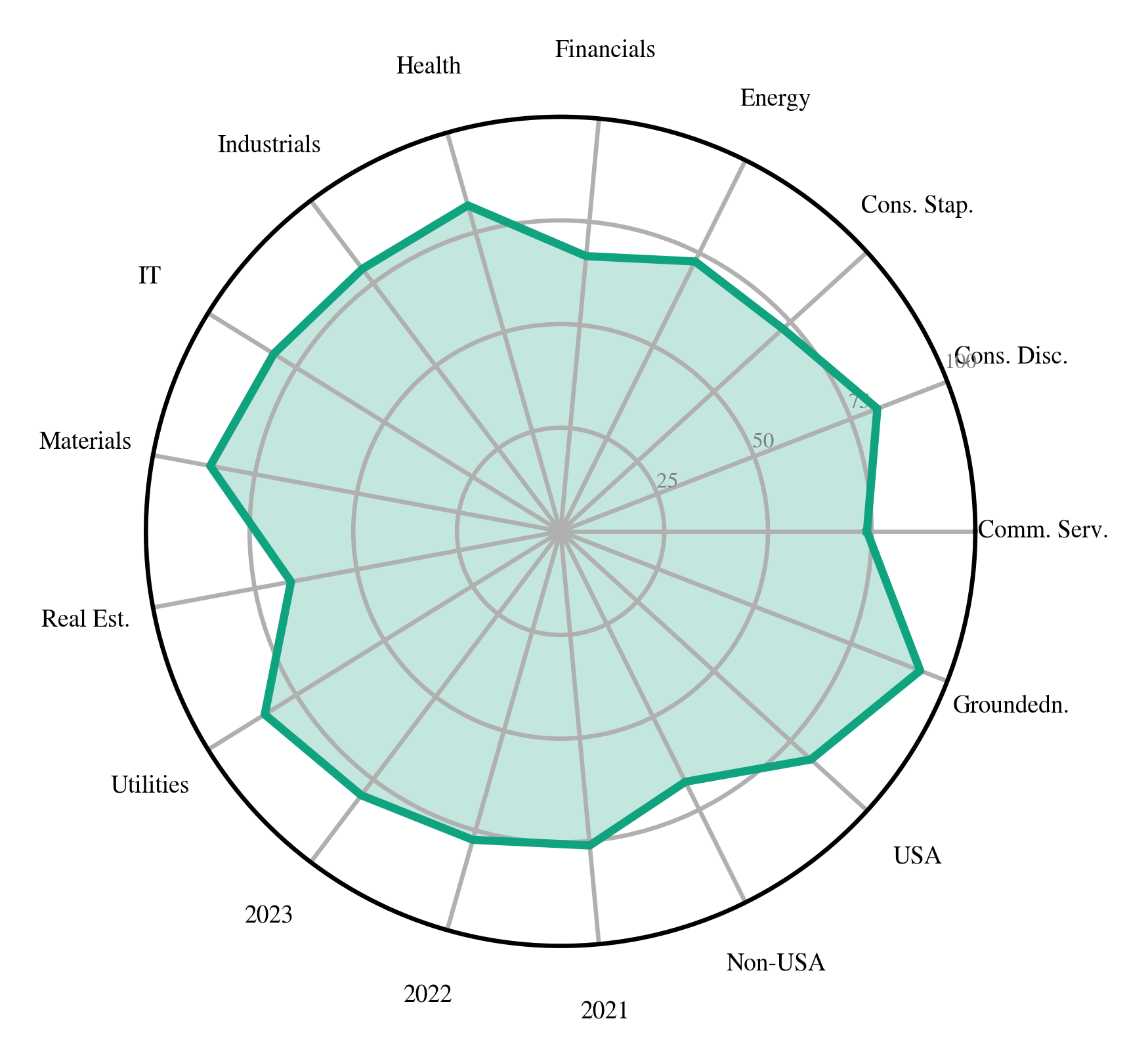}{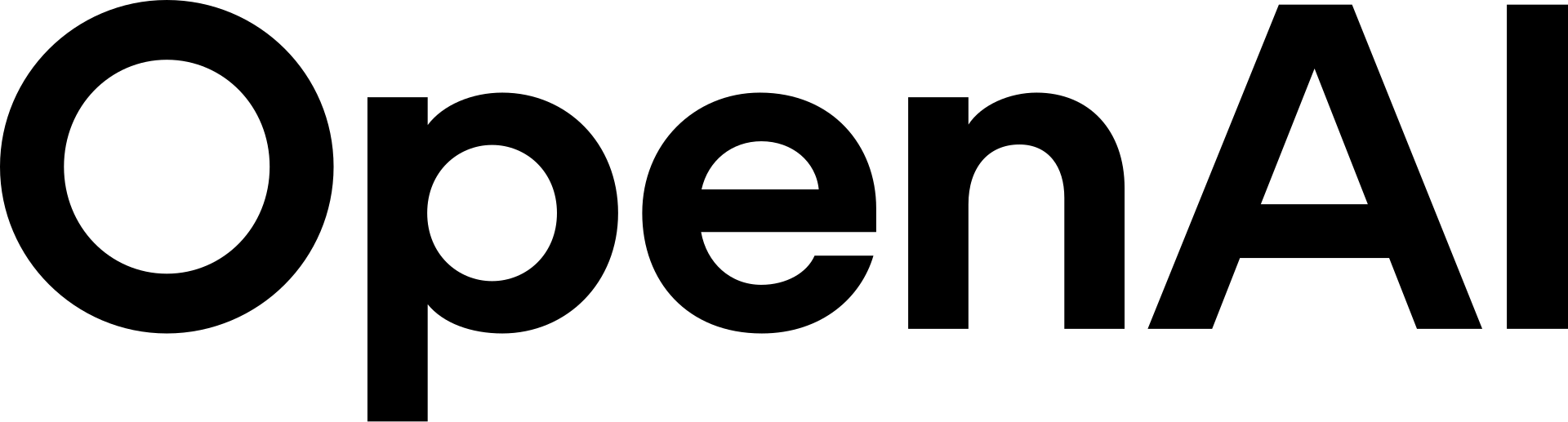}
                \caption{o4-mi.\ }
                \vspace{-0.1cm}
                \label{ft12:fig:spider_chart_o4-mini}
            \end{subfigure}
            \begin{subfigure}[b]{0.11\textwidth}
                \centering
                \addlogo{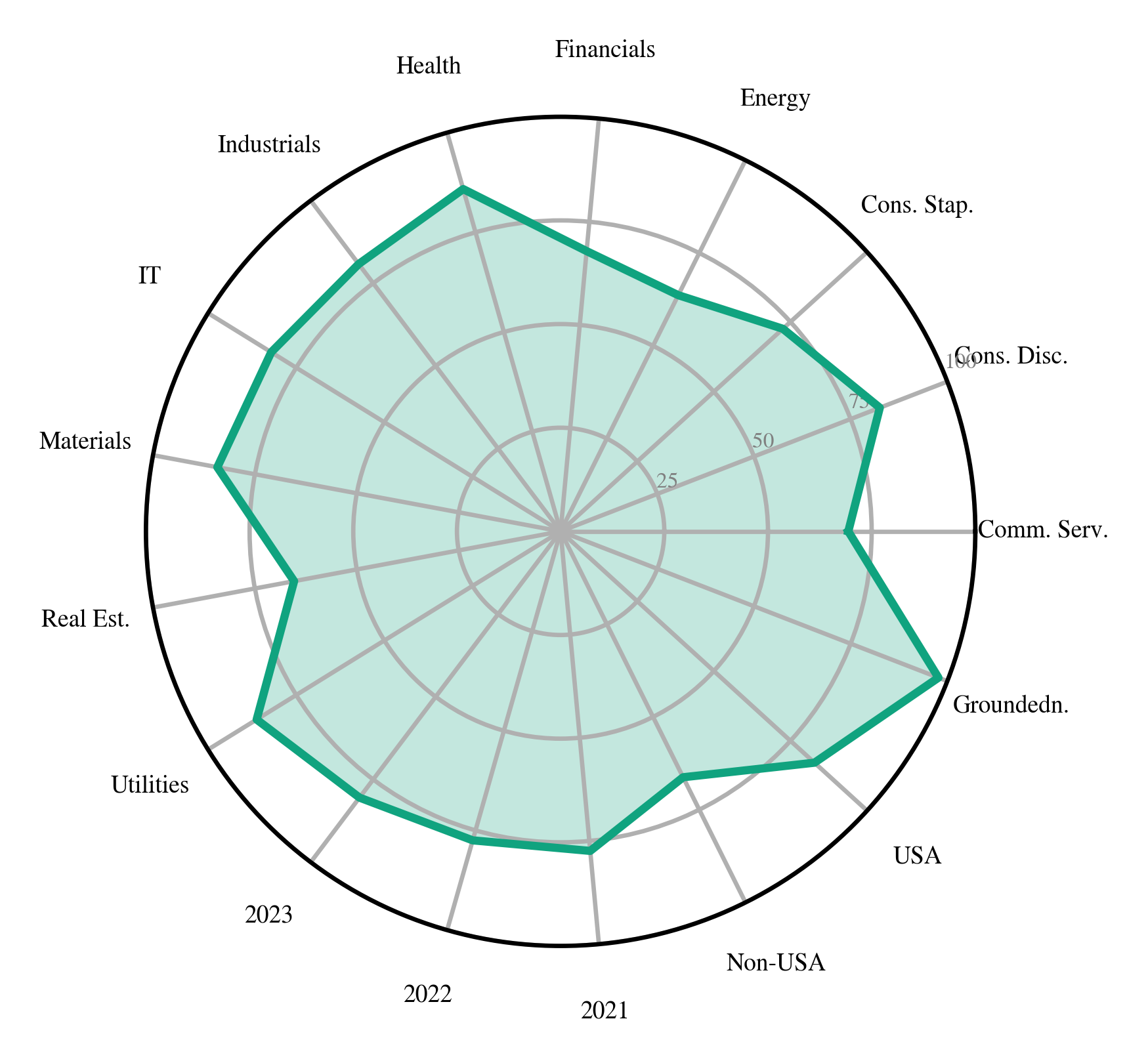}{images_final/logo_openai.png}
                \caption{" 5.2}
                \vspace{-0.1cm}
                \label{ft12:fig:spider_chart_gpt-5.2}
            \end{subfigure}
            \begin{subfigure}[b]{0.11\textwidth}
                \centering
                \addlogo{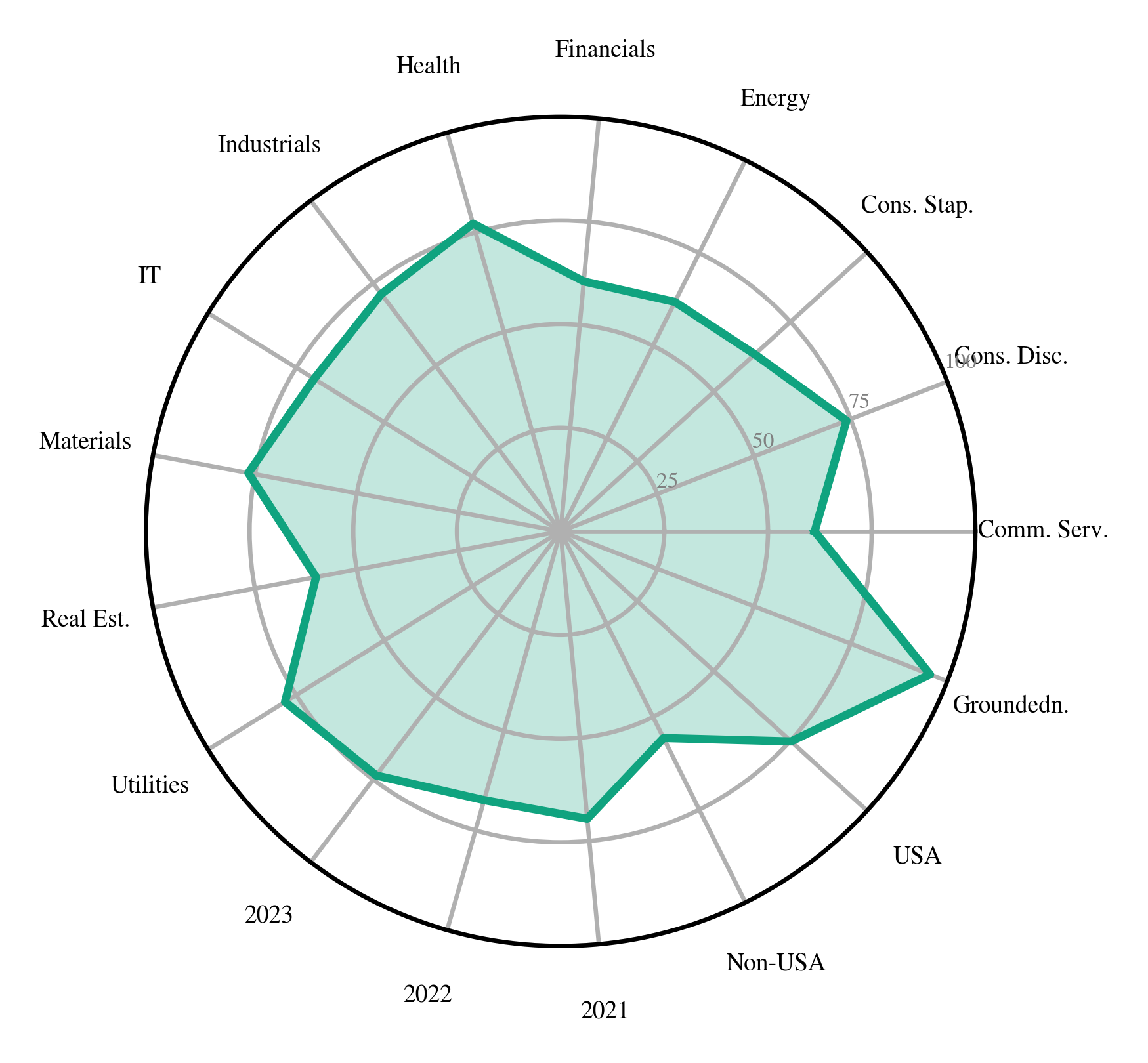}{images_final/logo_openai.png}
                \caption{" 4o}
                \vspace{-0.1cm}
                \label{ft12:fig:spider_chart_gpt-4o}
            \end{subfigure}

            \vspace{0.5em}

            %  Row 3: DeepSeek + xAI 
            \begin{subfigure}[b]{0.11\textwidth}
                \centering
                \addlogo[width=0.28\textwidth]{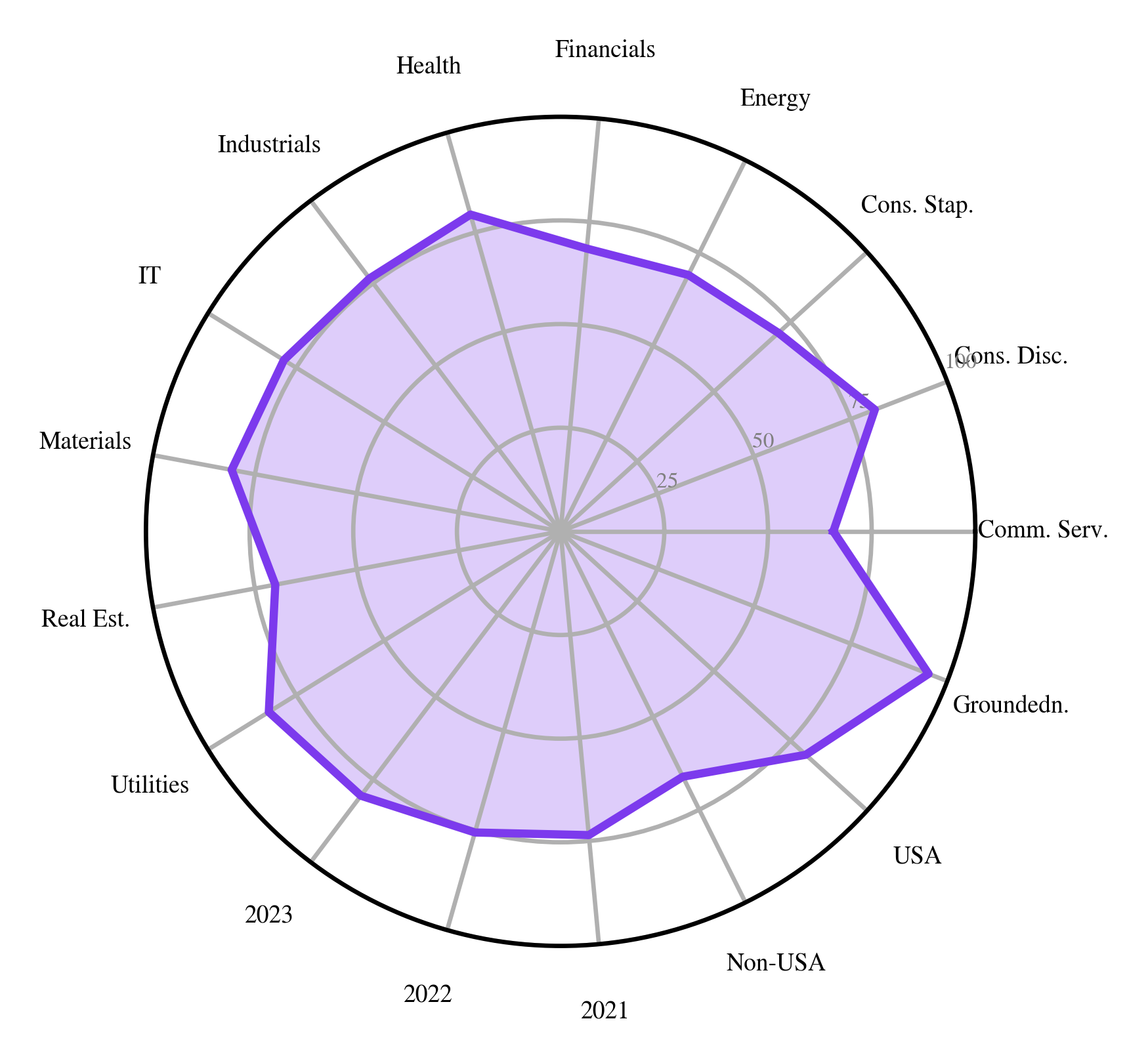}{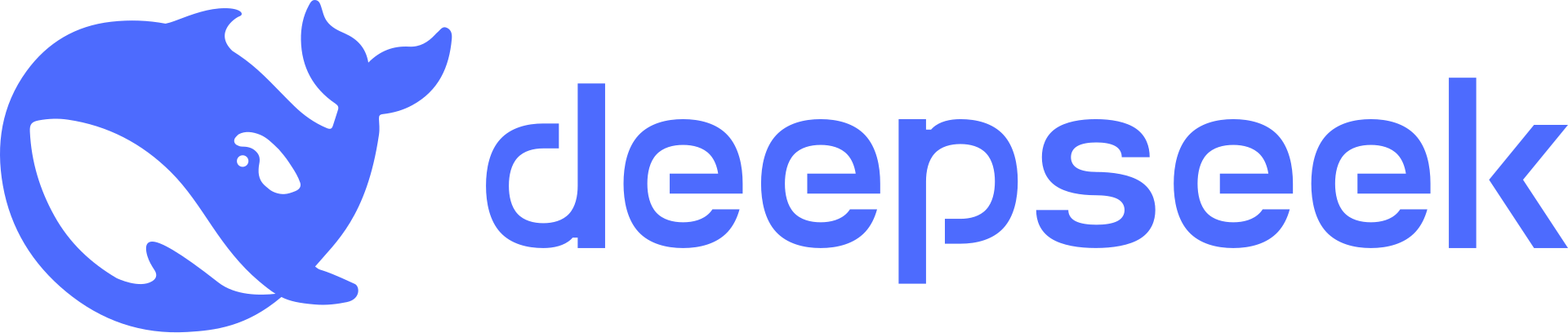}
                \caption{DS R1}
                \vspace{-0.1cm}
                \label{ft12:fig:spider_chart_deepseek-reasoner}
            \end{subfigure}
            \begin{subfigure}[b]{0.11\textwidth}
                \centering
                \addlogo[width=0.28\textwidth]{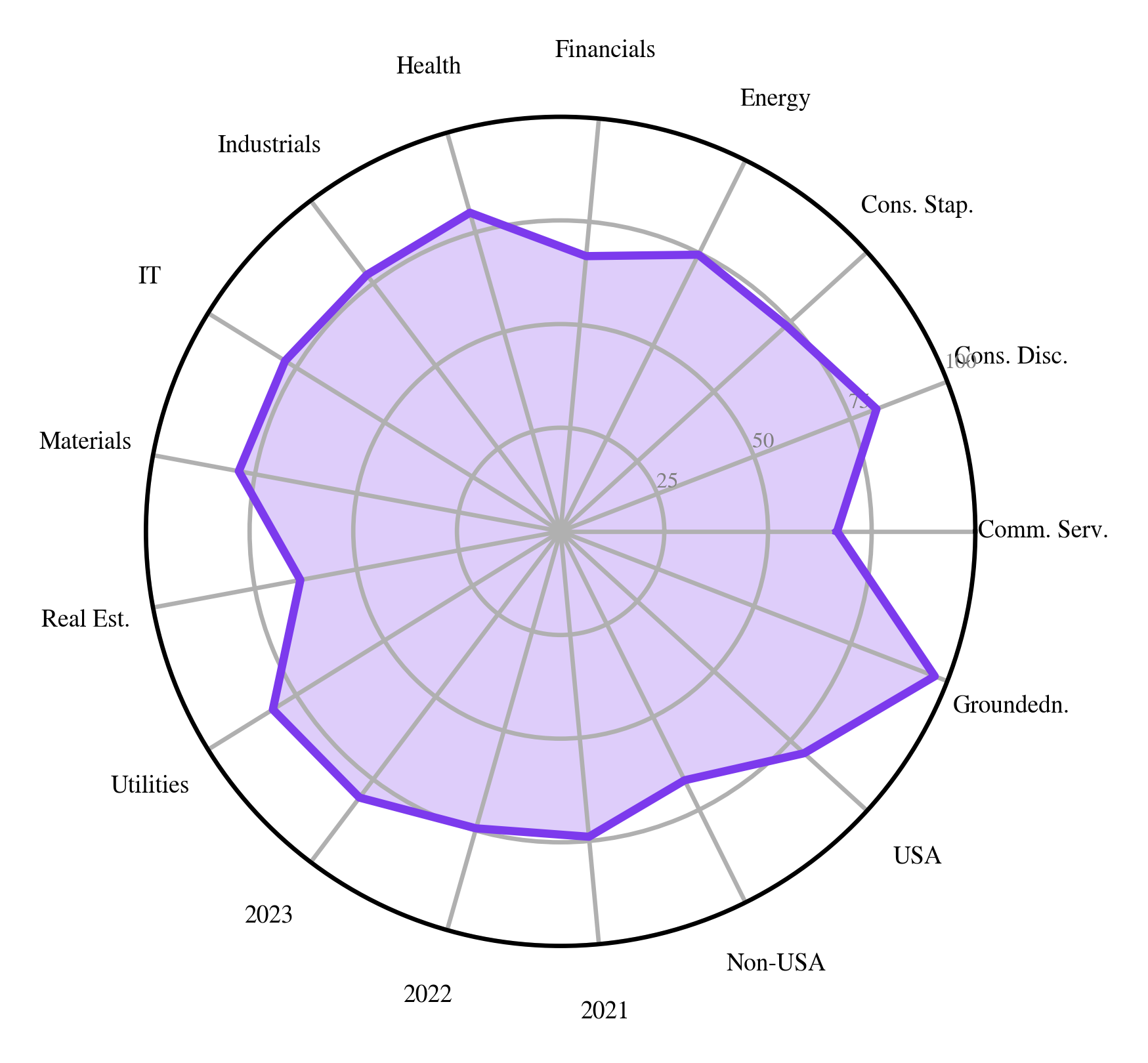}{images_final/logo_deepseek.png}
                \caption{" V3.1}
                \vspace{-0.1cm}
                \label{ft12:fig:spider_chart_deepseek-chat}
            \end{subfigure}
            \begin{subfigure}[b]{0.11\textwidth}
                \centering
                \addlogo[width=0.28\textwidth]{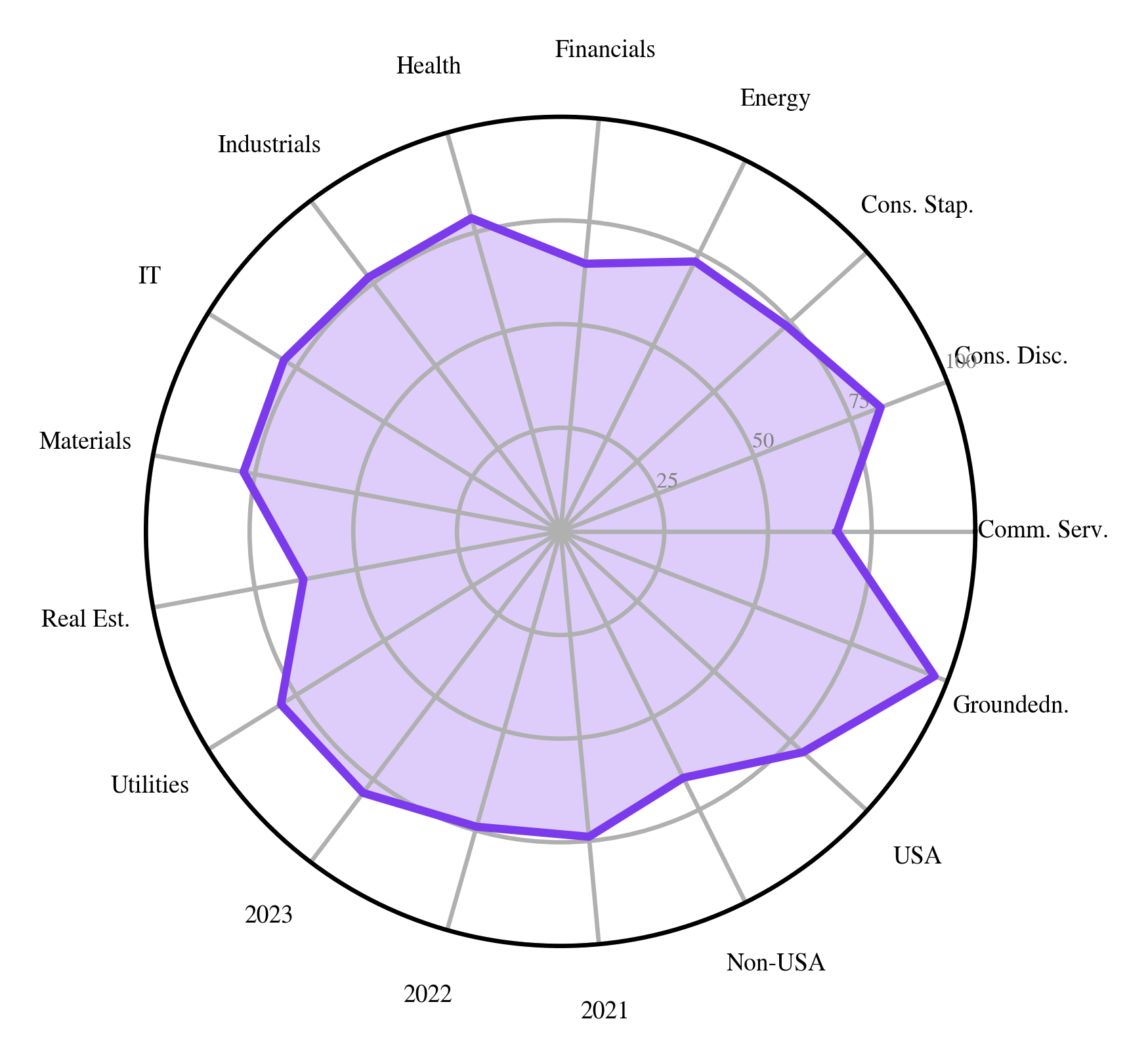}{images_final/logo_deepseek.png}
                \caption{" V3.2}
                \vspace{-0.1cm}
                \label{ft12:fig:spider_chart_deepseek-v3.2}
            \end{subfigure}
            \begin{subfigure}[b]{0.11\textwidth}
                \centering
                \addlogo[width=0.1\textwidth]{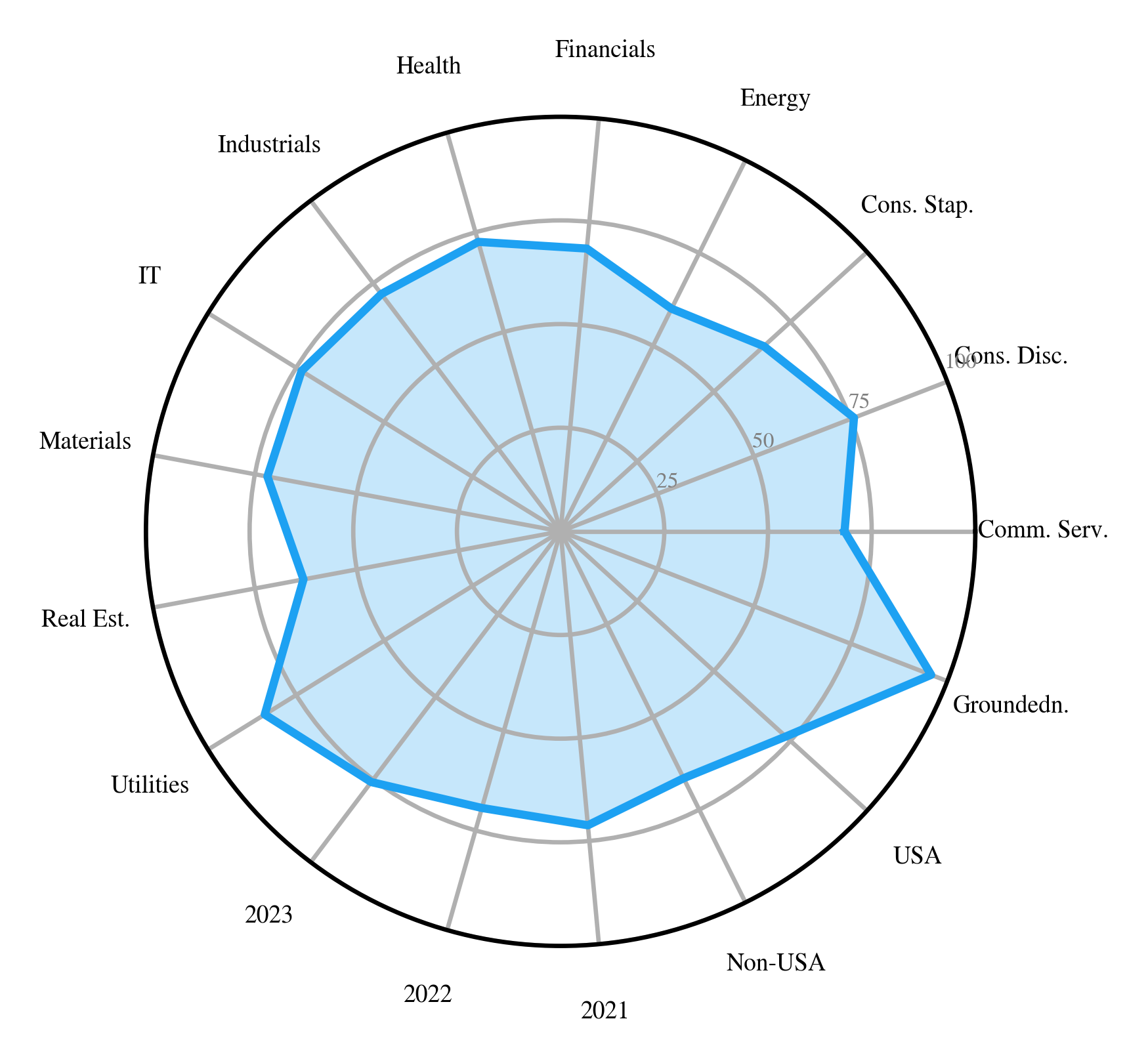}{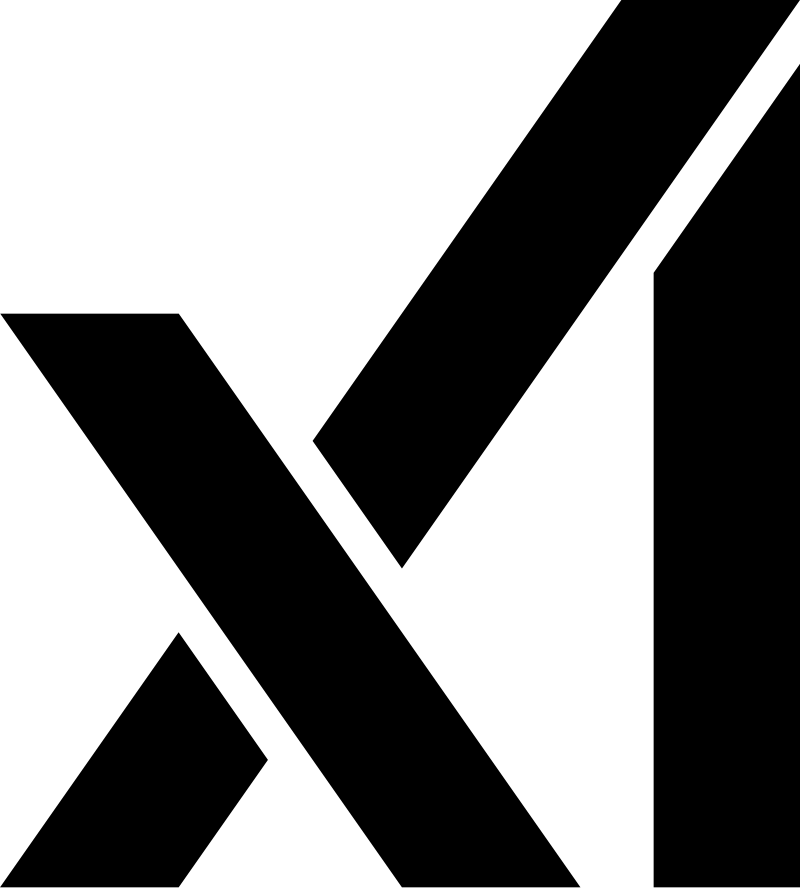}
                \caption{Grok 4}
                \vspace{-0.1cm}
                \label{ft12:fig:spider_chart_grok-4}
            \end{subfigure}

            \vspace{0.5em}

            % Row 4: xAI + Zhipu + Mistral 
            \begin{subfigure}[b]{0.11\textwidth}
                \centering
                \addlogo[width=0.1\textwidth]{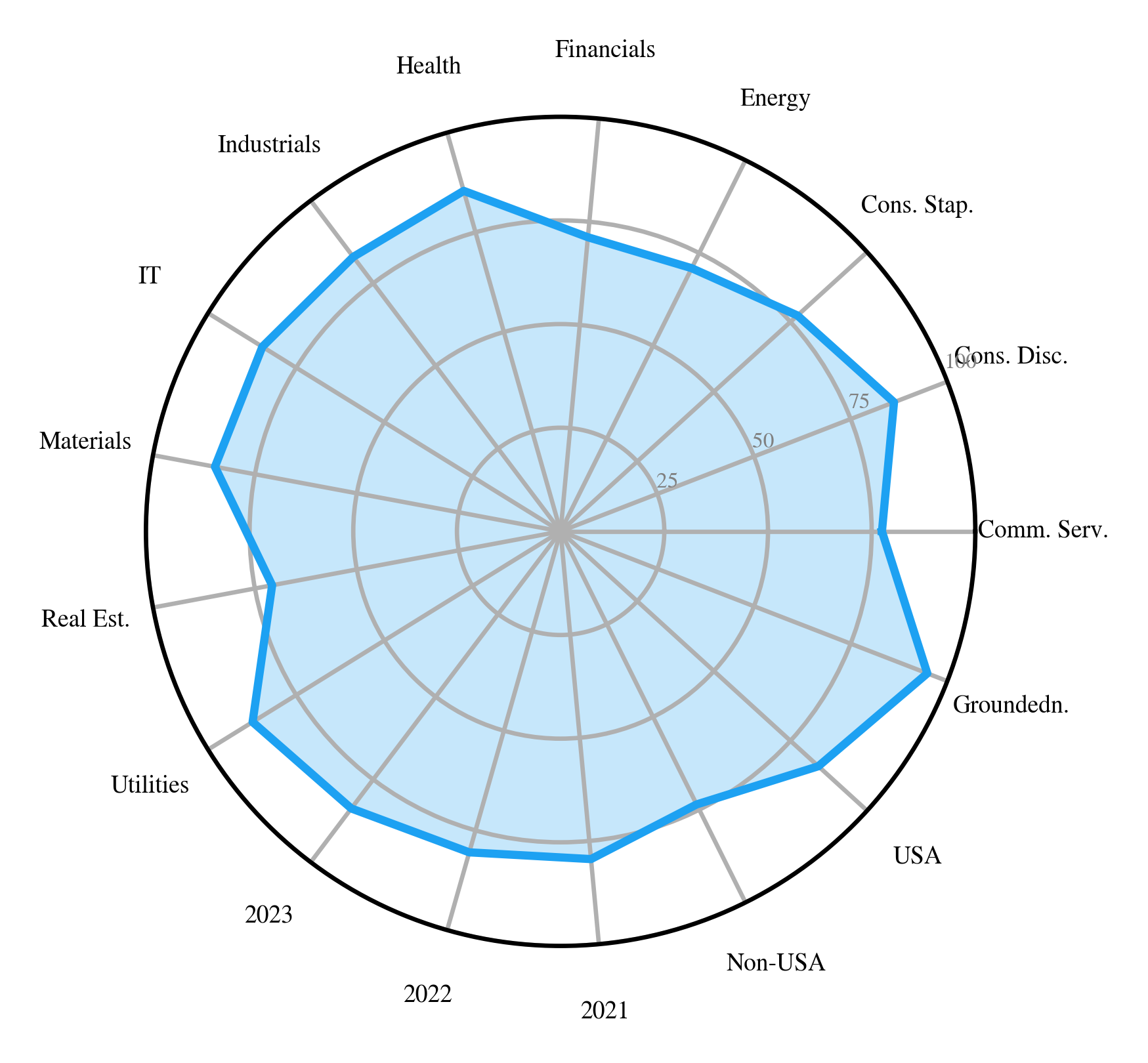}{images_final/logo_xai.png}
                \caption{" 4.1}
                \vspace{-0.1cm}
                \label{ft12:fig:spider_chart_grok-4.1}
            \end{subfigure}
            \begin{subfigure}[b]{0.11\textwidth}
                \centering
                \addlogo{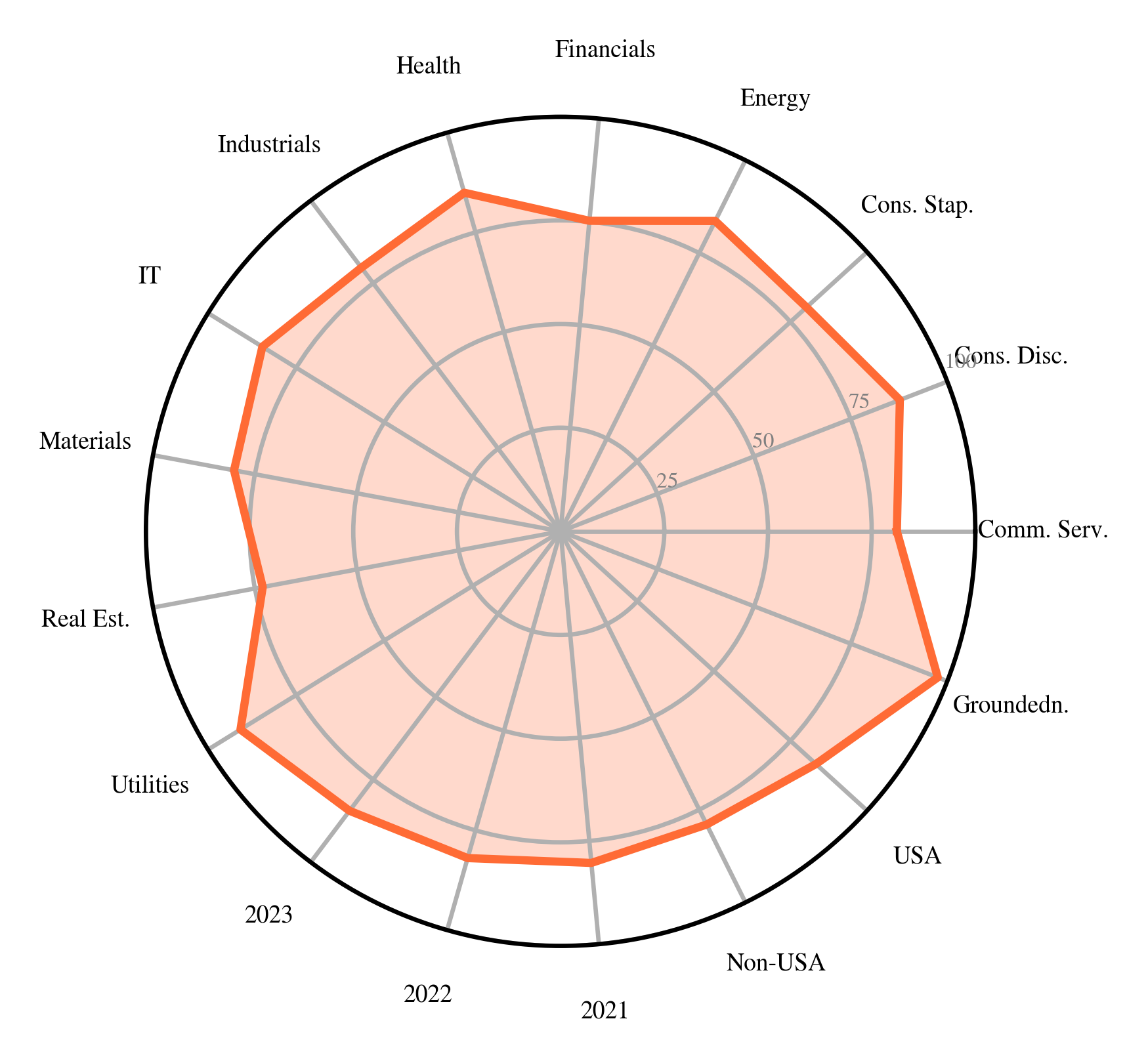}{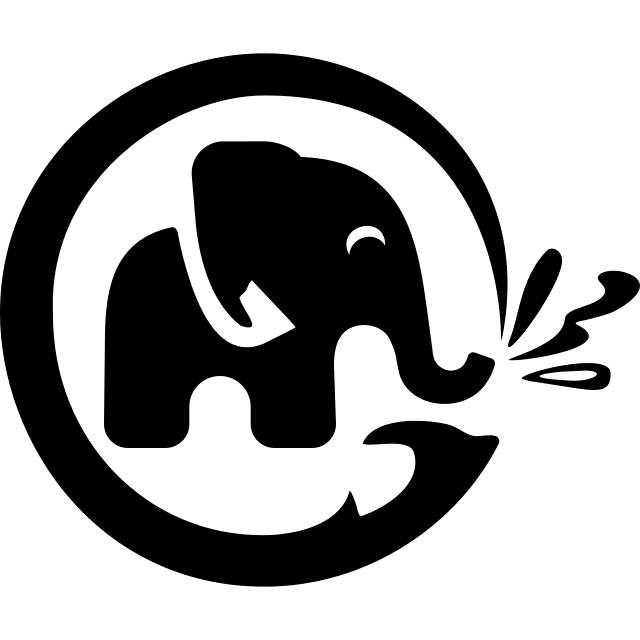}
                \caption{GLM 5}
                \vspace{-0.1cm}
                \label{ft12:fig:spider_chart_glm-5}
            \end{subfigure}
            \begin{subfigure}[b]{0.11\textwidth}
                \centering
                \addlogo{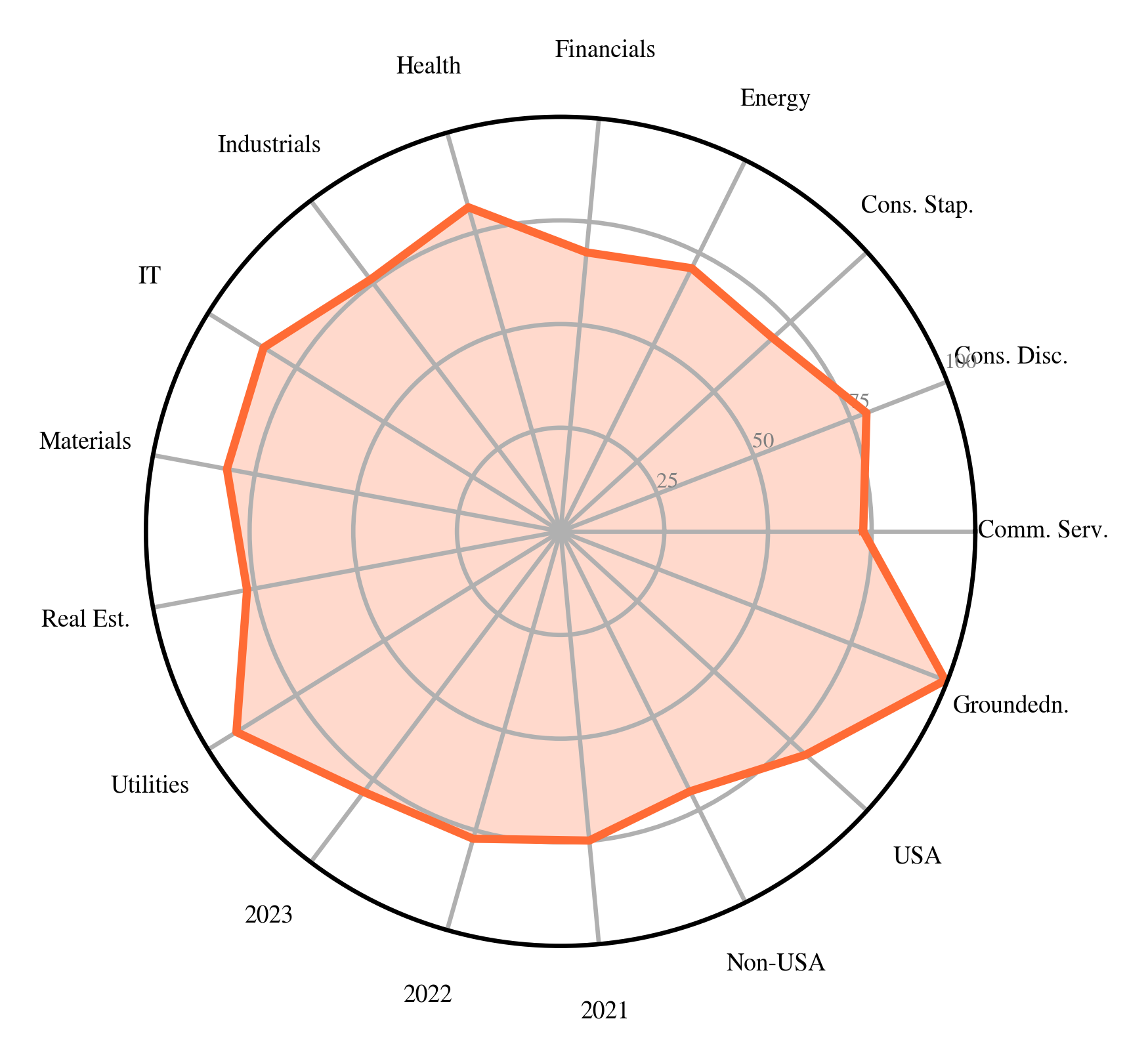}{images_final/logo_zhipu.png}
                \caption{" -4.7}
                \vspace{-0.1cm}
                \label{ft12:fig:spider_chart_glm-4.7-thinking}
            \end{subfigure}
            \begin{subfigure}[b]{0.11\textwidth}
                \centering
                \addlogo{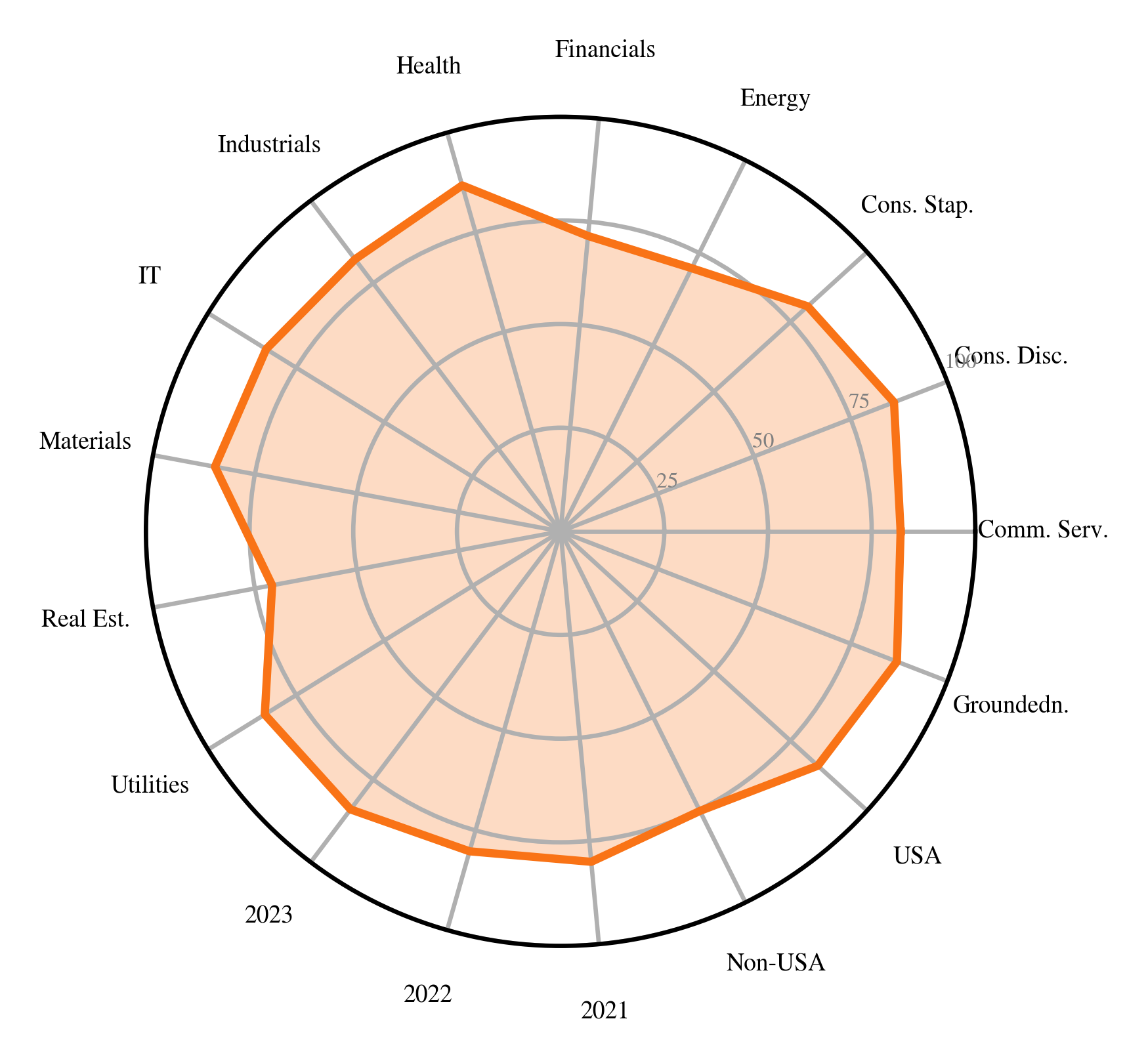}{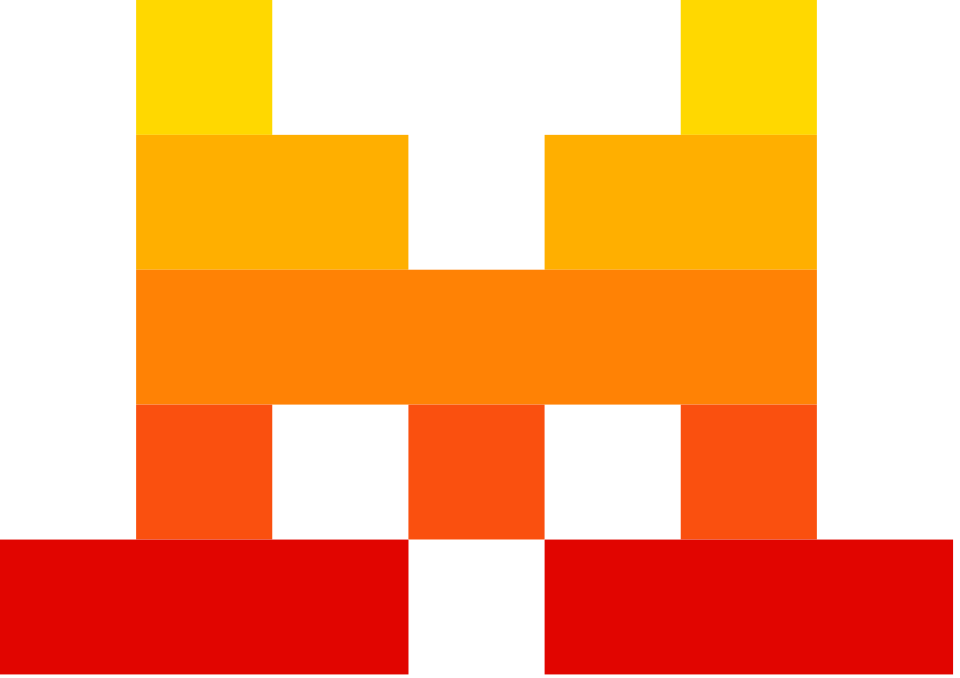}
                \caption{Mis.\ 3}
                \vspace{-0.1cm}
                \label{ft12:fig:spider_chart_mistral-3}
            \end{subfigure}

            \vspace{0.5em}

            % Row 5: Mistral + Alibaba + Moonshot + Baidu
            \begin{subfigure}[b]{0.11\textwidth}
                \centering
                \addlogo{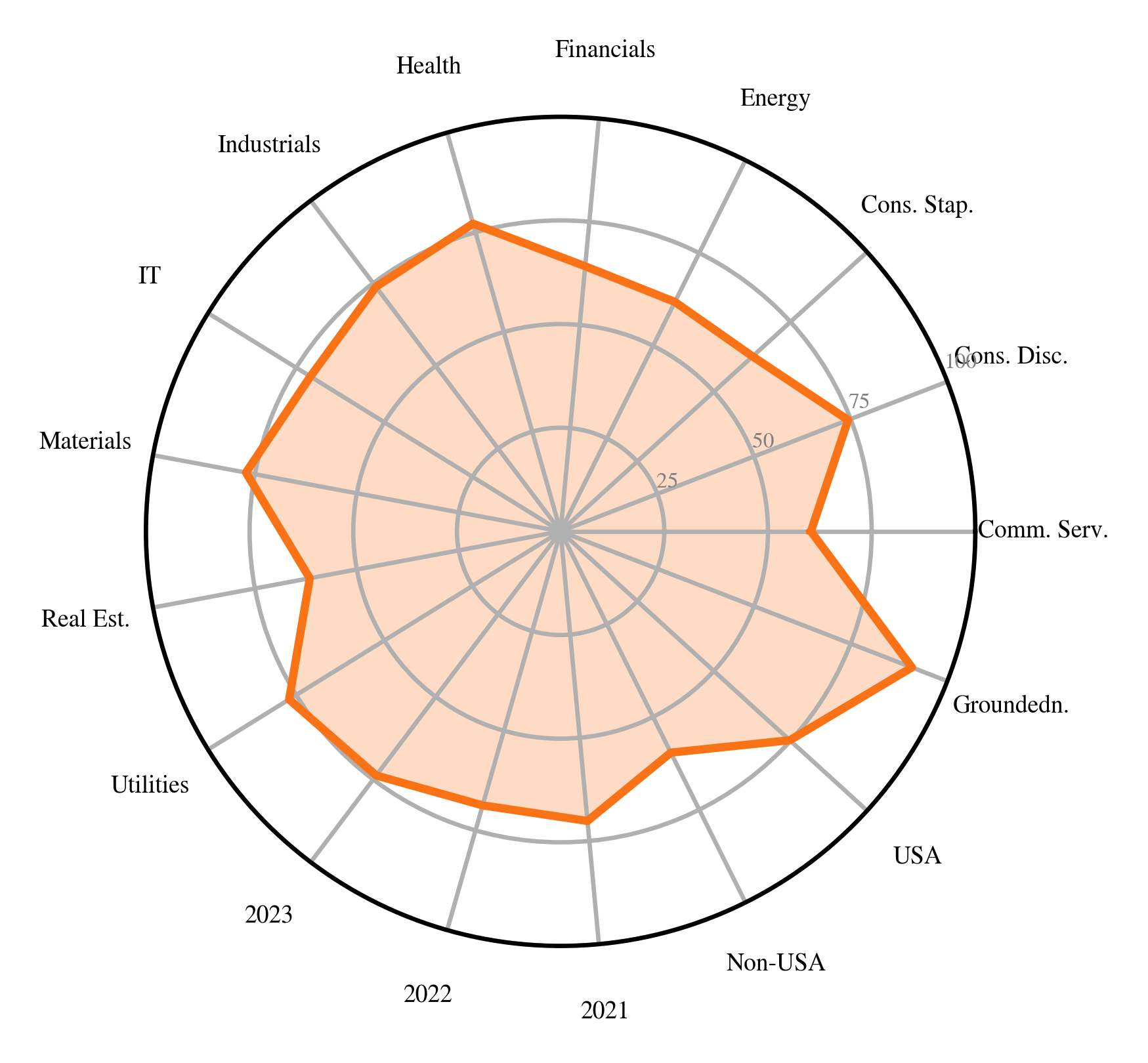}{images_final/logo_mistral.png}
                \caption{" Magi.}
                \vspace{-0.1cm}
                \label{ft12:fig:spider_chart_magistral}
            \end{subfigure}
            \begin{subfigure}[b]{0.11\textwidth}
                \centering
                \addlogo{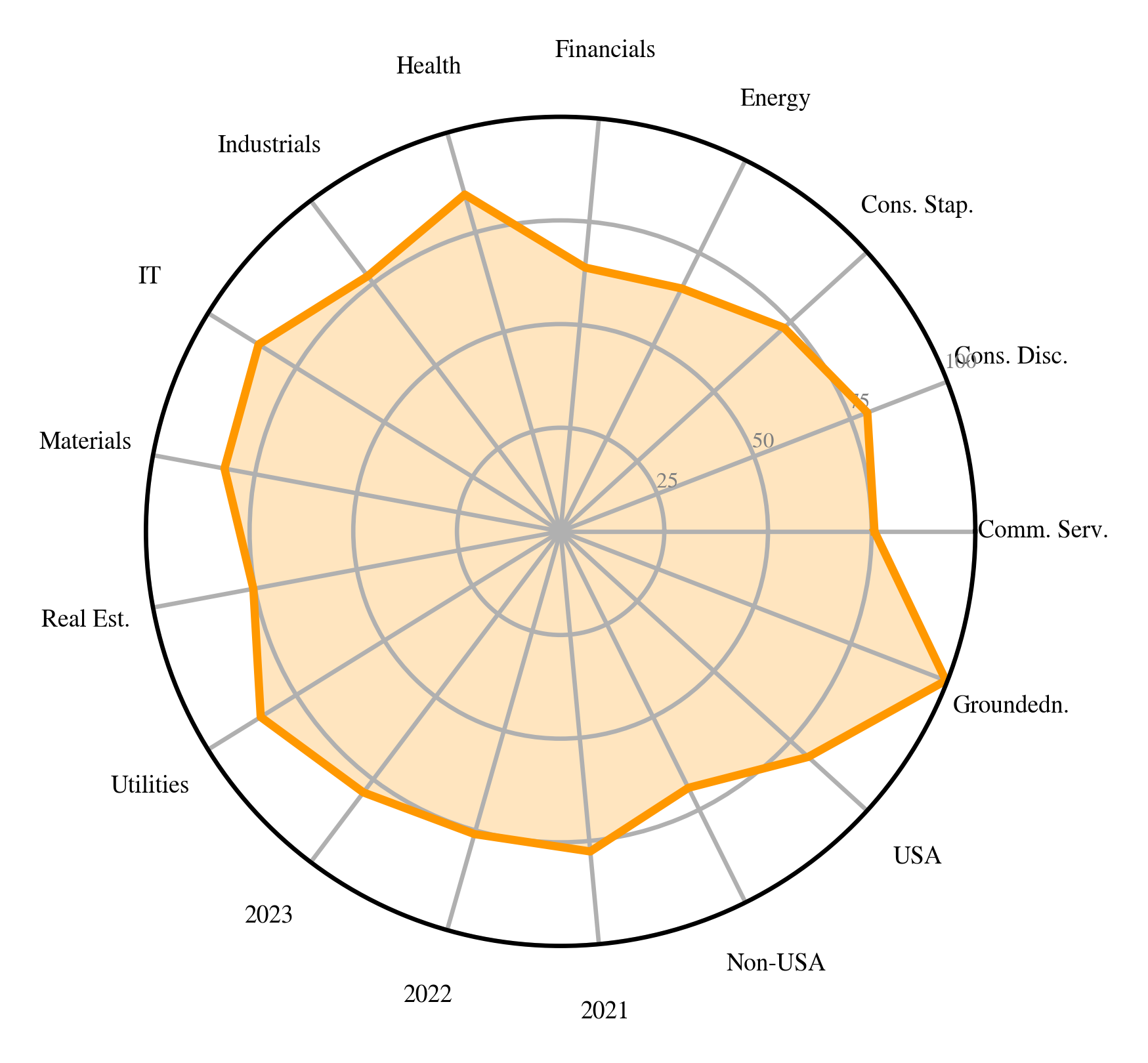}{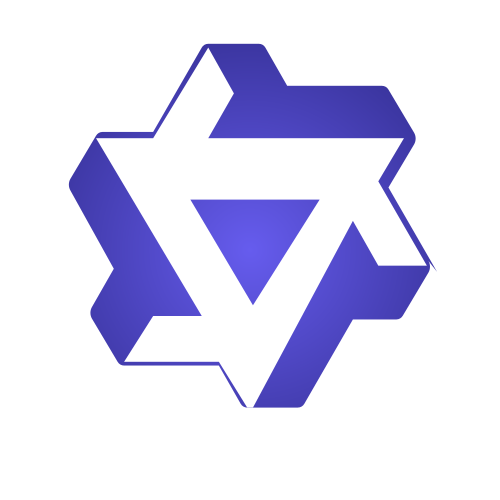}
                \caption{Qw3-M.}
                \vspace{-0.1cm}
                \label{ft12:fig:spider_chart_qwen-3}
            \end{subfigure}
            \begin{subfigure}[b]{0.11\textwidth}
                \centering
                \addlogo{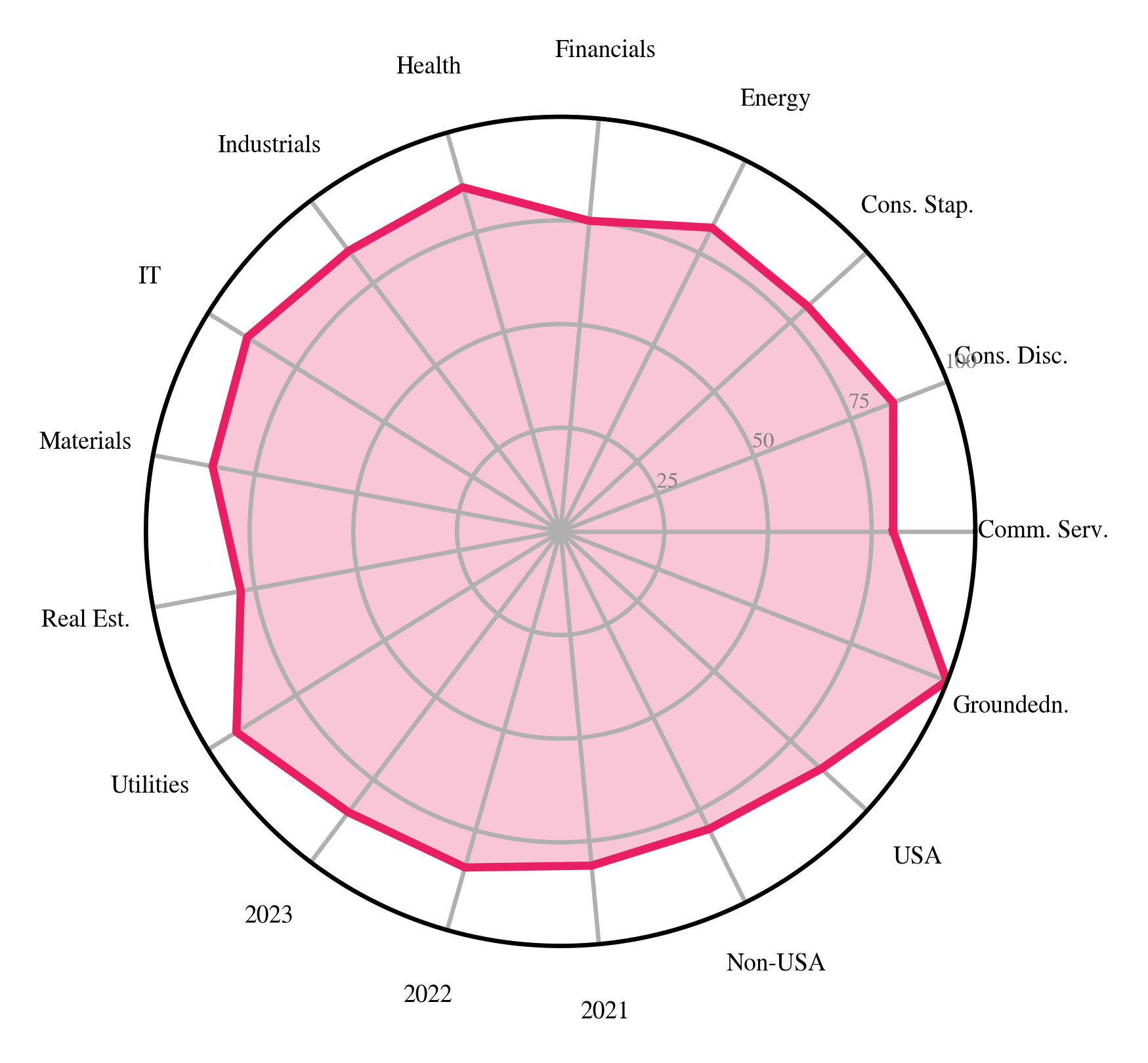}{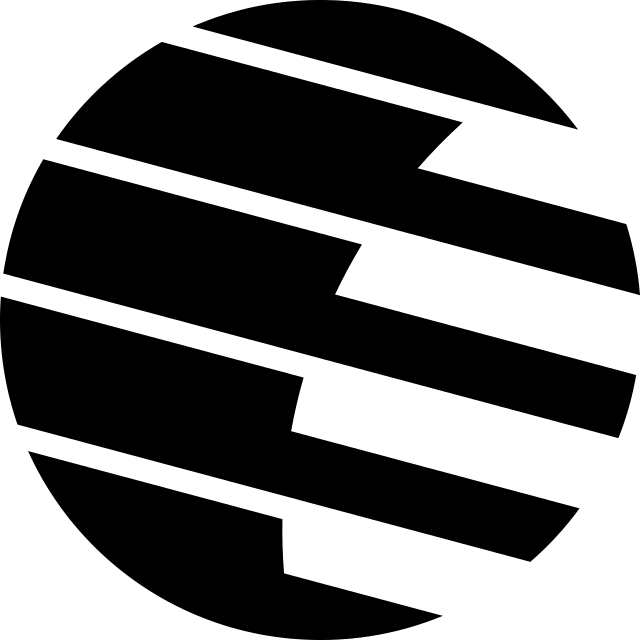}
                \caption{K.\ K2.6}
                \vspace{-0.1cm}
                \label{ft12:fig:spider_chart_kimi-k2.5}
            \end{subfigure}
            \begin{subfigure}[b]{0.11\textwidth}
                \centering
                \addlogo{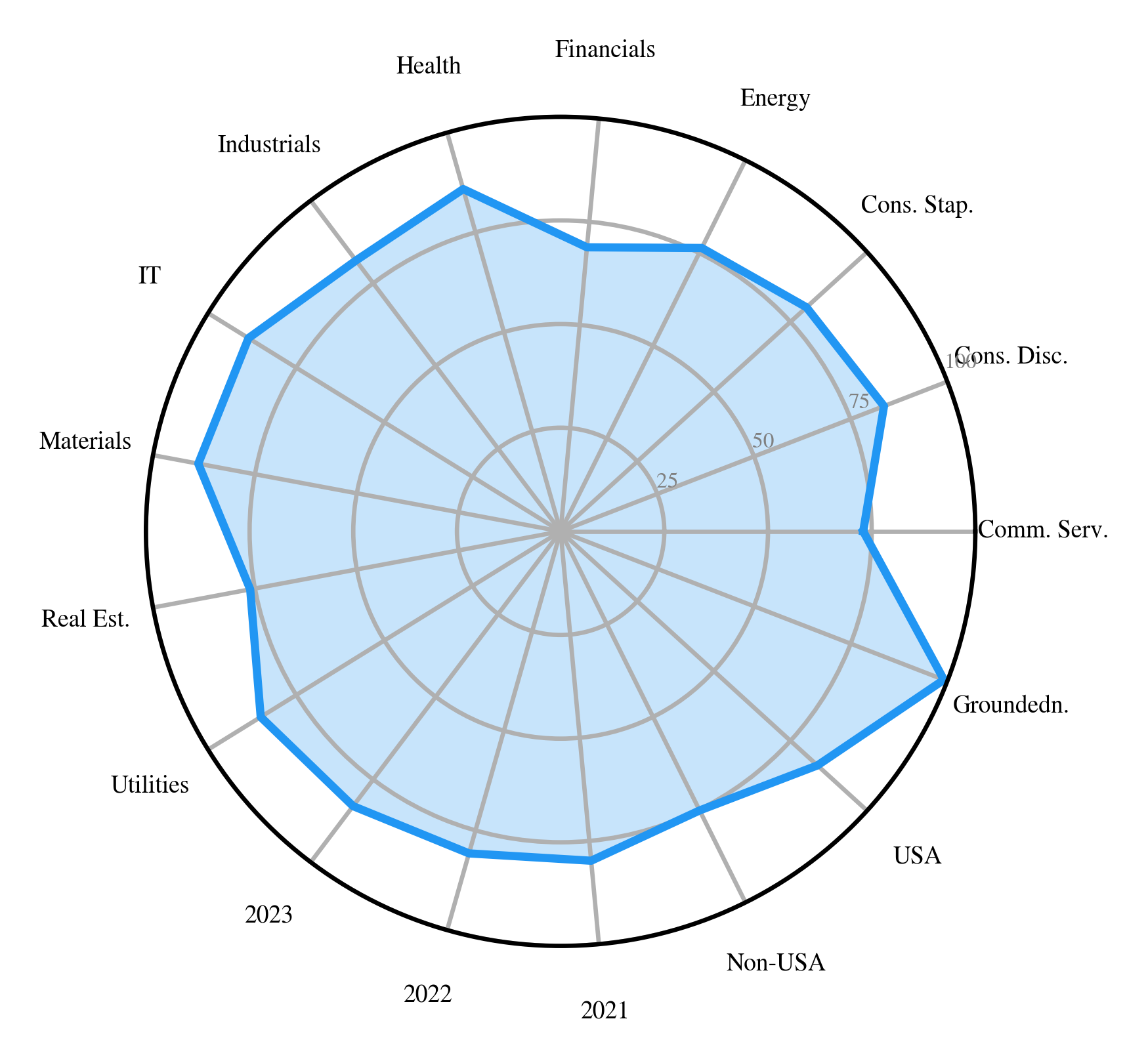}{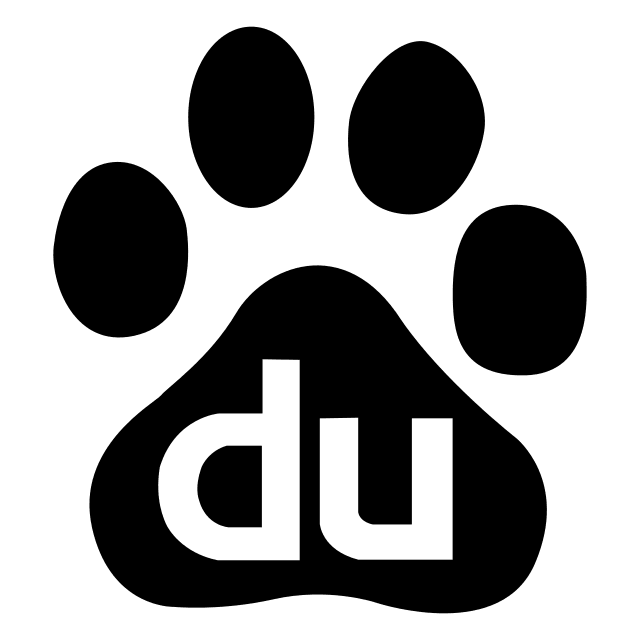}
                \caption{ER.\ 5.1}
                \vspace{-0.1cm}
                \label{ft12:fig:spider_chart_ernie-5.0}
            \end{subfigure}

            \vspace{0.5em}

            %  Legend 
            \begin{subfigure}[b]{0.22\textwidth}
                \centering
                \parbox{\linewidth}{
                    \centering
                    \scriptsize
                    \setlength{\tabcolsep}{1pt}

                    \begin{tabular}{@{}ll@{}}
                        \cmidrule(r){1-2}
                        \multicolumn{2}{l}{\textcolor{colorIndustries}{\textit{--- Industries ---}}} \\
                        \textcolor{colorIndustries}{1. Comm. Services}    & \textcolor{colorIndustries}{7. Industr.} \\
                        \textcolor{colorIndustries}{2. Consumer Discr.}   & \textcolor{colorIndustries}{8. IT} \\
                        \textcolor{colorIndustries}{3. Consumer Stapl.}  & \textcolor{colorIndustries}{9. Materials} \\
                        \textcolor{colorIndustries}{4. Energy}            & \textcolor{colorIndustries}{10. Real Est.} \\
                        \textcolor{colorIndustries}{5. Financials}        & \textcolor{colorIndustries}{11. Utilities} \\
                        \textcolor{colorIndustries}{6. Health}            & \\
                    \end{tabular}

                    \begin{tabular}{@{}ll@{}}
                        \cmidrule(r){1-2}
                        \multicolumn{2}{l}{\textit{--- Other Metrics ---}} \\
                        \textcolor{colorYears}{12. 2023}             & \textcolor{colorOther}{15. Non-USA} \\
                        \textcolor{colorYears}{13. 2022}             & \textcolor{colorOther}{16. USA} \\
                        \textcolor{colorYears}{14. 2021}             & \textcolor{colorOther}{17. Groundedn.} \\
                        \cmidrule(r){1-2}
                    \end{tabular}
                }
                \caption{Legend}
                \label{ft12:fig:spider_legend}
            \end{subfigure}

        \end{figure}
    
    \subsection{Model Performance Comparison}

        We employ multiple evaluation metrics. First, accuracy, measured as recall: does the answer state the required information correctly and completely? Second, the hallucination rate, i.e., the share of claims that the source documents do not support; groundedness, its complement, corresponds to the precision of the system. Third, a failure taxonomy that attributes every error either to the retrieval stage or to the generating LLM. 
        % Fourth,  we provide test statistics to describe the confidence of the results. 
        As indicated in the caption of figure~\ref{ft12:fig:performance_barchart_sufficient_retrieval.png}, the model results exclude retrieval errors, which isolates the models' pure comprehension performance. The human baseline involves no retrieval stage; its error rate is therefore an end-to-end figure.

        Claude Opus 4.6 is the best performing model by accuracy (88.4\%), closely followed by Claude Sonnet 4.6 (86.7\%). Moonshot AI's open-weight Kimi~K2.6 ranks third (83.5\%); all three surpass the human accuracy baseline of 82.8\%. GLM~5 (82.0\%) and Mistral~3 (81.3\%) round out the top five by accuracy, though Mistral~3 has a notably high hallucination rate of 13.0\%.

        The ranking by hallucination rate tells a different story. Gemini 2.5 Pro has a near-zero hallucination rate (0.08\%), below the human baseline of approx.\ 2.8\%, but its accuracy (76.6\%) is only thirteenth-best. Kimi~K2.6 comes close (0.13\%) while ranking third in accuracy, making it the strongest combination of accuracy and groundedness in the field. Three further Chinese models follow in hallucination prevention: Qwen3-Max (0.38\%), GLM-4.7 (0.54\%), and ERNIE~5.1 (0.88\%), all ahead of Gemini~3~Pro (1.7\%) and GPT-5.2 (2.3\%).

        Grok~4.1 (80.9\% accuracy, 5.2\% hallucination) performs well on both dimensions. The three DeepSeek models cluster together at 75--76\% accuracy with moderate hallucination rates of 3--5\%. Among them, the reasoning model DeepSeek~R1 has higher accuracy but also more hallucinations than the non-reasoning V3.1 and V3.2.

        At the bottom of the ranking are GPT-4o (69.2\% accuracy, 4.5\% hallucination), Magistral (70.1\%, 9.2\%), and Grok~4 (71.3\%, 4.2\%). With two reasoning models among the bottom three, reasoning capability alone does not predict performance on this task; we return to this observation in the discussion.

    \subsection{Error Analysis}
 
        \subsubsection{Failure Taxonomy and Analysis}

        Our analysis of 20,682 failures across twenty models reveals that retrieval issues account for 48.9\% of all failures, followed by model comprehension errors (44.7\%) and standalone hallucinations (9.1\%). Note that the model error share has increased relative to the earlier eleven-model evaluation because the expanded set includes more models with weaker comprehension capabilities.

        \begin{table}%[h]
            \begin{center}

            \caption{\textbf{Failure Mode Distribution.} Model errors and hallucinations are not mutually exclusive.}
            \label{ft12:tab:failure_taxonomy}
            \tiny
            \begin{tabular}{@{}llr@{}}
                
                \toprule
                \textbf{Failure Type} & \textbf{Description} & \textbf{Rate} \\
                \midrule
                Retrieval Failure & Missing/partial information & 48.9\% \\
                Model Error & Comprehension failures & 44.7\% \\
                Hallucination & Fabricated information & 9.1\% \\
                \bottomrule
            \end{tabular}

            \end{center}
        \end{table}

        \subsubsection{Key Insights}
        
        \textbf{Imperfect performance correlation}: The correlation between accuracy and hallucination rate is not uniform, as figure~\ref{ft12:fig:performance_barchart_sufficient_retrieval.png} illustrates.
        \textbf{Cascading failures}: Poor retrieval triggers downstream hallucinations.
        \textbf{Hallucination protection varies widely}: Between Gemini 2.5 Pro (0.08\%) and Mistral~3 (13.0\%), there is a difference of about two orders of magnitude in the hallucination rate.

    \subsection{Illustrative Case Studies}

        \paragraph{Case Study 1: Near-Universal Difficulty in Segment Identification.}
        A question about Unilever's business segments from its 2021 annual report was answered correctly by only 2 of 20 models (GLM-4.7 and Qwen3-Max). The golden answer listed five strategic growth spaces (hygiene, skin care, prestige beauty, functional nutrition, plant-based foods), but the retrieved context also contained the three reporting segments (Beauty \& Personal Care, Foods \& Refreshment, Home Care). Models consistently confused strategic focus areas with the formal reporting segments, producing answers that were plausible but factually incomplete. This demonstrates that even when the retriever successfully provides the relevant context, models struggle with questions requiring disambiguation between multiple valid segmentation dimensions within a single document.

        \paragraph{Case Study 2: Straightforward Company Type Extraction.}
        In contrast, a question about Toyota's legal form of incorporation from its 2021 annual report was answered correctly by all twenty models. The answer (a Japanese \emph{Kabushiki Kaisha}, or joint-stock company) was stated clearly in the report. This case confirms that when the answer is unambiguous and prominently placed, models perform reliably across the board.

        % Old version:
        % \paragraph{Case Study 2: Straightforward Company Type Extraction.}
        % In contrast, a question about Toyota's legal form of incorporation from its 2021 annual report was answered correctly by 19 of 20 models. The answer (a Japanese \emph{Kabushiki Kaisha}, or joint-stock company) was stated clearly in the report. Only Kimi~K2.5 failed this question, hallucinating an incorrect legal form. This case confirms that when the answer is unambiguous and prominently placed, nearly all models perform reliably.

        \paragraph{Case Study 3: Hallucination Hotspot on Indian Reports.}
        A question about revenue growth from Bajaj Finance's 2023 annual report (India) was answered correctly by only 2 of 20 models, while 6 models hallucinated. The Indian report format, which presents financial data across multiple tables with varying fiscal year conventions, confused most models. Claude Opus~4.6, Claude Sonnet~4.6, GPT-5.2, and Grok~4.1 all hallucinated revenue growth figures that appeared plausible but were not supported by the retrieved context. This pattern is consistent with the broader finding that Indian reports produce the highest hallucination rates.

        \paragraph{Case Study 4: Cross-Lab Divergence on Indian Revenue.}
        A question about Tata Consultancy Services' revenue from its 2022 annual report (India) was answered correctly by 12 of 20 models but produced a striking pattern: GLM~5, Qwen3-Max, and DeepSeek~V3.2 answered correctly, while Claude Opus~4.6 and GPT-5.2 did not. This suggests that for certain non-US report formats, Chinese models may have an advantage, potentially due to broader coverage of Asian financial documents in their training data. The question required extracting revenue from a table formatted in Indian \emph{crores} (tens of millions), a unit unfamiliar to most Western-trained models.

    \subsection{Content Filter Refusals in Chinese Models}\label{ft12:subsection:chinese-model-refusals}

        An important finding from our evaluation of Chinese open-weight models is the presence of content filters that cause refusals of purely financial questions. Across 23,736 answer attempts from the eight Chinese-provider models, 18 refusals persisted in the benchmark data (0.08\%). No non-Chinese model refused any question. Refusal behavior is, moreover, a property of the model \emph{and} the access route: Kimi~K2.6, called via the native Moonshot API, refused 20 questions as ``high risk,'' yet the identical requests, routed through a third-party proxy to the same vendor backend, were all answered. Table~\ref{ft12:tab:chinese_refusals} summarizes the affected models, routes, and companies.

        The primary trigger is the mention of Chinese political leaders in the retrieved context. All three DeepSeek models refused the same five questions about Kweichow Moutai, a major Chinese state-owned company whose annual report quotes a speech by General Secretary Xi Jinping. The DeepSeek content filter returned a ``Content Exists Risk'' error, despite the underlying questions being purely financial (revenue growth, segments, key financials).

        The 20 native-route Kimi~K2.6 refusals reproduce and extend this trigger taxonomy: eight concern Kweichow Moutai (the same political-leader trigger that affects DeepSeek), four concern Industrial Bank, a Chinese state-owned bank, two concern Ralph Lauren (see below), and six concern Nintendo, a Japanese video game company whose annual report contains no politically sensitive content; these six refusals are apparent false positives. Because the proxy route answered all 20 questions, the benchmark stores the answered responses; Kimi~K2.6's accuracy is unaffected, and the native-route refusals are reported as a property of the endpoint.

        References to Hong Kong protests triggered refusals even in reports from non-Chinese companies. Ralph Lauren's US SEC Form 10-K mentions ``adverse impacts related to COVID-19 and Hong Kong protest business disruptions,'' which was sufficient to trigger both the GLM~5 filter (error code 1301, Zhipu's native content-filter rejection) and the native Moonshot filter. GLM~5 additionally refused one question about PTT Public Company (Thailand) without a discernible trigger.

        GLM-4.7, Qwen3-Max, and ERNIE~5.1 produced no refusals at all, but all three were accessed through proxy routes (OpenRouter and NanoGPT; table~\ref{ft12:tab:model-snapshots}). The Kimi~K2.6 contrast shows that a zero refusal count observed on a proxy route does not establish that the vendor's native endpoint would not filter; refusal rates are only meaningful relative to the exact endpoint used.

        While the overall impact on our benchmark is small (0.08\% of attempts), these findings reveal a systematic constraint on the deployment of Chinese AI models for global financial analysis, where annual reports routinely reference jurisdictions and political figures that may trigger content filters. They also carry a methodological lesson for benchmark design: refusal behavior must be attributed to model--route pairs rather than to models alone, since a proxy can silently mask a vendor's content filter.

        \begin{table}[h]
            \centering
            \caption{\textbf{Content-Filter Refusals Are a Property of Model and Access Route.} Refusals observed across the eight Chinese-provider models (23{,}736 attempts). 18 refusals persisted in the benchmark data (0.08\%). Kimi~K2.6 additionally refused 20 questions when called via the native Moonshot API but answered all 20 when the identical requests were routed through the NanoGPT proxy; the benchmark stores the answered responses. Unmarked routes are the vendors' native APIs; * marks third-party proxy routes. Zero counts on proxy-only routes are therefore route-conditional and do not imply that the native endpoints would not filter. No non-Chinese model refused any question.}
            \label{ft12:tab:chinese_refusals}
            \tiny
            \setlength{\tabcolsep}{4pt}
            \begin{tabular}{@{}llr>{\raggedright\arraybackslash}p{2.9cm}@{}}
                \toprule
                \textbf{Model} & \textbf{Route} & \textbf{Ref.} & \textbf{Companies} \\
                \midrule
                Kimi~K2.6      & Moonshot    & 20 & Kweichow Moutai (CN), Nintendo (JP), Industrial Bank (CN), Ralph Lauren (US) \\
                Kimi~K2.6      & NanoGPT*    &  0 & same 20 questions, all answered \\
                DeepSeek R1    & DeepSeek    &  5 & Kweichow Moutai (CN) \\
                DeepSeek V3.1  & DeepSeek    &  5 & Kweichow Moutai (CN) \\
                DeepSeek V3.2  & DeepSeek    &  5 & Kweichow Moutai (CN) \\
                GLM 5          & Zhipu       &  3 & Ralph Lauren (US), PTT Public (TH) \\
                \midrule
                GLM-4.7        & OpenRouter* &  0 & --- \\
                Qwen3-Max      & OpenRouter* &  0 & --- \\
                ERNIE 5.1      & NanoGPT*    &  0 & --- \\
                \bottomrule
            \end{tabular}
        \end{table}

    \subsection{Addressing Potential Sources of Experimental Bias in Favor of Selected Models}

        Gemini 2.5 Pro's lead in hallucination prevention (0.08\%) raises the question of whether experimental conditions unfairly favor this model. We do not think so: while the Gemini model family has the largest context window of all models under consideration, we only utilize less than 10\% of it. If we increased the number of tokens retrieved from annual reports, we would expect Gemini to widen its lead as some models' context windows would approach their limits. Notably, Gemini 2.5 Pro ranks only thirteenth in accuracy, indicating that its hallucination advantage does not translate into overall dominance. Also, Kimi~K2.6 is close to Gemini with a 0.13\% hallucination rate.

        \begin{figure}%[t]
            \begin{center}
                \caption{\textbf{Average Performance by Question Type.} Correct and hallucination rates are end-to-end shares of all questions of a type, averaged across the twenty models; they include the questions with failed retrieval and are therefore not conditional on retrieval success. The retriever-error rate is the share of questions whose top-5 retrieved chunks were insufficient; it is a property of the shared retrieval pipeline and identical for all models. The three bars are separate rates and do not sum to 100\%.}
                \includegraphics[width=7.5cm]{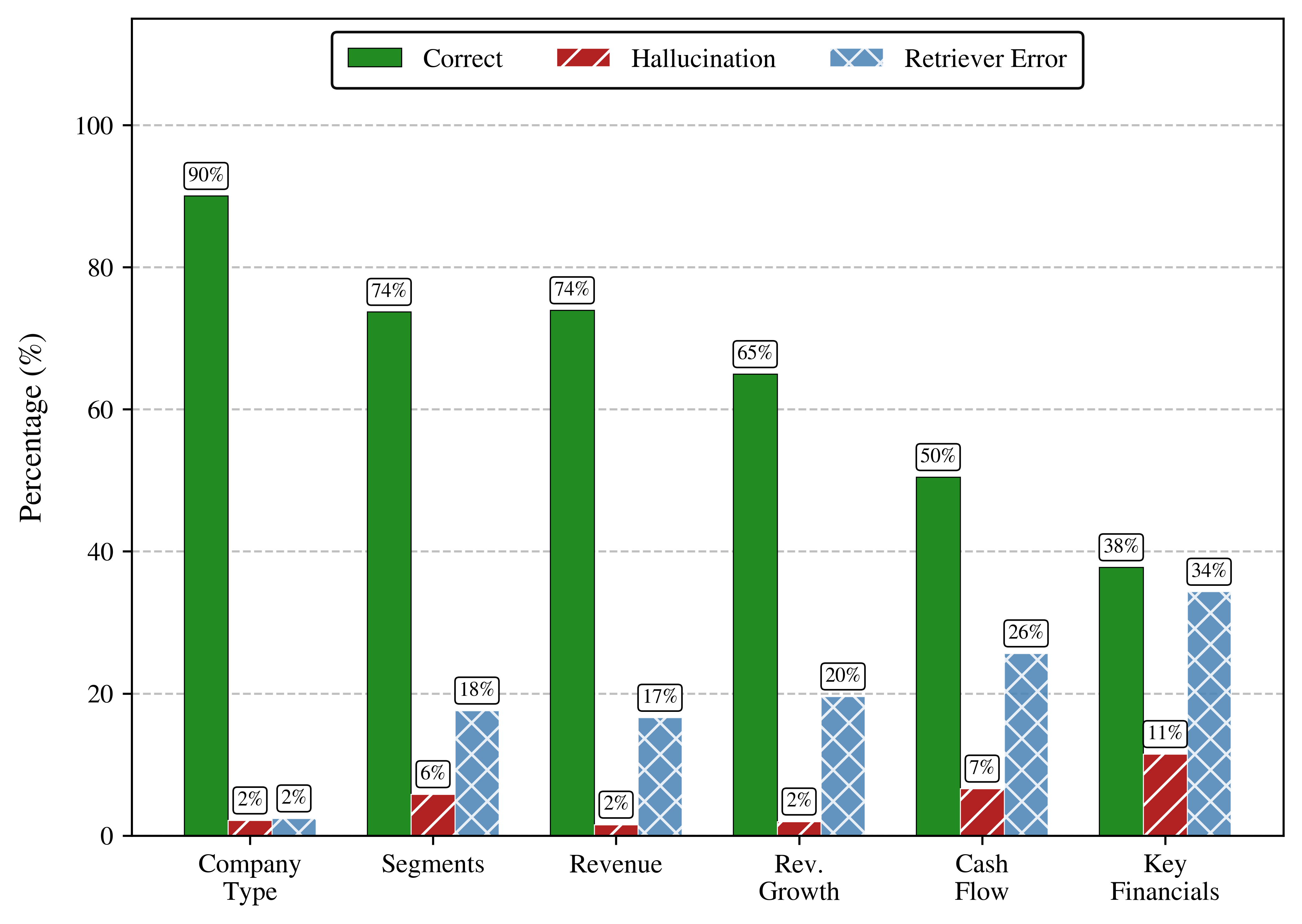}
                % % \Description{Average Performance by Question Type: The question about key financials is by far the hardest question for the retriever to answer, while the cash flow question frequently leads to wrong model results.}
                \label{ft12:fig:accuracy_by_question.png}
            \end{center}
            % \Description{The question about key financials is by far the hardest question for the retriever to answer, while the cash flow question frequently leads to wrong model results.}
        \end{figure}

    \subsection{Checks for Training Data Contamination}\label{ft12:subsection:results-checks-for-training-data-contamination}

        Because our model set includes several models released after \cite{spoerer_2025} was written, we need to re-examine whether any of them could have been trained on our golden answers in the meantime. We approach this question in three ways: an empirical stress test, a geographic sanity check, and a temporal robustness check.

        \paragraph{Retrieval Dependence as a Contamination Probe.}
        The most direct evidence against contamination is how strongly accuracy depends on the retrieved context. Pooled across all twenty models, accuracy is 77.9\% when the retrieved chunks are sufficient but 12.0\% on the 575 questions where retrieval fails, a more than sixfold drop. A model that had memorized our golden answers would be largely insensitive to this difference. The residual 12\% is not itself evidence of memorization: an insufficient-retrieval label means the judge deemed the chunks incomplete for the full golden answer, not devoid of information, and partial context sometimes carries the headline figure. Conditional accuracy under retrieval failure does vary, from 6.6\% (Grok~4) through 12--13\% for the two Claude models to 20--29\% for the four models evaluated last (Kimi~K2.6, ERNIE~5.1, GLM-4.7, Qwen3-Max). We read the top of this range primarily as stronger extraction from partial context, but note that these four models combine the latest training cut-offs with a different judge version (Appendix~\ref{ft12:app:judge}), so reasoning, judge behavior, and residual leakage cannot be fully separated for them; the geographic and temporal checks below show no corresponding anomaly.

        \paragraph{Geographic Sanity Check.}
        A second contamination signature would be an uneven distribution of model strength across jurisdictions. A model trained more heavily on US SEC filings would outperform its competitors specifically on US reports. Figure~\ref{ft12:fig:usa_vs_row_geo_performance_barchart_sufficient_retrieval.png} shows the opposite: the ranking of models is broadly stable across US and non-US reports, with all models performing slightly better on US filings (a difficulty effect, not a memorization effect). There is no evidence of idiosyncratic model advantages on a specific country's reports.

        \paragraph{Temporal Robustness Check.}
        If older reports had leaked into training data but newer reports had not, we would expect a monotonic decline in accuracy as report year advances. What we actually observe is flat-to-slightly-improving accuracy: 76.8\%, 77.4\%, and 81.0\% average accuracy (excluding retriever errors) in 2021, 2022, and 2023, respectively. We take this as further evidence that the models are reading the retrieved context rather than recalling internalized financial figures.

        \paragraph{Theoretical Plausibility.}
        Finally, it is worth noting how little of the training corpus an annual-report benchmark would occupy. Our data has about 83 million tokens, or about 0.001\% of a typical frontier-model training budget. For a model to memorize year-specific company financials would require the training pipeline to weight our specific reports unusually heavily, which we have no reason to expect. We deliberately stratified the dataset to include both large and lesser-known companies precisely to make memorization less likely.

        %\begin{figure}[h]
        %    \ \\
        %    \begin{center}
        %    
        %        \caption{\textbf{Performance by Annual Report Year}: Performance does not significantly vary by year.}
        %        \includegraphics[width=8.3cm]{images_final/year_performance_barchart_sufficient_retrieval.png}
        %        \label{ft12:fig:year_performance_barchart_sufficient_retrieval.png}
        %
        %    \end{center}
        %\end{figure}

        % No  bias as a result of training set contamination
        % We looked for signs of report data leaking into the models' training data by comparing the average accuracy for each report year in figure ~\ref{ft12:fig:year_performance_barchart_sufficient_retrieval.png}. Stronger performance on older reports would indicate that some models were already trained on these older reports, which could complicate performance comparisons 

        % Acknowledge that this is not a 100% guarantee
        While this analysis cannot rule out that a leakage effect is present, we wanted to report this finding for full transparency. In our view, the demonstrated retrieval dependence bounds the possible impact of leakage: when the correct context is missing, the models consistently fail, which suggests that the answers were not recalled from parametric (trained) world knowledge.% There is a high possibility that some annual reports, especially the older ones, leaked into some of the datasets used by the twenty models under consideration.

        %\begin{figure}[t]
        %    \begin{center}
        %        \caption{\textbf{Average Performance by Industry, Excluding Retriever Errors.}}
        %        % \Description{Average Performance by Industry, Excluding Retriever Errors: There are moderate differences in difficulty across industries.}
        %        \includegraphics[width=8.3cm]{images_final/industry_performance_barchart_sufficient_retrieval.png}
        %        \label{ft12:fig:industry_performance_barchart_sufficient_retrieval.png}
        %    \end{center}
        %\end{figure}

\section{Discussion and Conclusion}

Running largely the same benchmark across twenty frontier models instead of eleven changes the picture that emerged from \cite{spoerer_2025} in three substantive ways.

\paragraph{Claude, Not Gemini, Tops the Accuracy Leaderboard.}
The prior version of this benchmark was headlined by Gemini~2.5 Pro, which dominated both accuracy and hallucination prevention on the eleven-model set. With the expanded set, Claude Opus~4.6 and Claude Sonnet~4.6 take the top two accuracy slots, each surpassing the 82.8\% human accuracy baseline. Gemini~2.5~Pro retains its distinctive advantage in one dimension only: hallucination prevention, where its 0.08\% rate is the lowest, well under the 2.8\% human rate, and far ahead of even Gemini~3~Pro (1.7\%). When retrieval failures are included, Kimi~K2.6 (0.13\%) even edges out Gemini~2.5~Pro (0.17\%) for the lowest hallucination rate overall.

\paragraph{Open-Weight and Non-Reasoning Models Have Closed the Gap.}
In the previous evaluation, reasoning capability was a clean predictor of accuracy: reasoning models dominated, non-reasoning models trailed. That story no longer holds. The strongest non-reasoning model, Mistral~AI's Mistral~3 (fifth), ranks above every reasoning model except Claude~Opus~4.6, Claude~Sonnet~4.6, Moonshot~AI's Kimi~K2.6 (third), and GLM~5 (fourth), beating frontier reasoning models such as GPT-5.2, Gemini~3~Pro, o4-mini, and DeepSeek~R1. Conversely, Magistral, Mistral~AI's dedicated reasoning model, ranks only 19th, fourteen places below its non-reasoning sibling Mistral~3, and Grok~4 (18th) trails three of the four non-reasoning models in the field.

\paragraph{Content Filters Are a New, Measurable Deployment Risk.}
A finding that has no analog in \cite{spoerer_2025} is the empirical impact of content filters in Chinese open-weight models. Across 23,736 answer attempts from the eight Chinese-provider models we tested, 18 refusals persisted in the benchmark data, and Kimi~K2.6's native endpoint refused 20 further questions that a proxy route answered. The refusals were triggered by Xi~Jinping mentions, a Chinese state-owned bank, Hong Kong protest references, or for no discernible reason. The effect is small in aggregate (0.08\%) but concentrated: the DeepSeek models uniformly refuse the same Kweichow~Moutai questions. It is also route-dependent, as identical requests that the native Moonshot API rejected as high-risk were all answered through a third-party proxy.

\paragraph{The Retrieval Bottleneck Persists.}
Across all twenty models, accuracy drops from 77.9\% to 12.0\% when the retriever fails to surface sufficient context. Retrieval failures account for 48.9\% of all errors in the expanded set. A mid-ranked model paired with a strong retriever may outperform the best model paired with our baseline retriever, and improvements to the RAG stack offer more headroom than swapping generators.

\paragraph{Scope and Future Work.}
Three scoping decisions limit the reach of our findings. First, all reports are in English; cross-lingual extension to Chinese, Japanese, and European-language filings is the clearest next step, especially given the content-filter dynamics we document here. Second, our six question types cover a representative but narrow slice of what an equity analyst actually asks; scaling toward the 169 question archetypes identified by \cite{popSpoerer_ffaq_identificationOfTheFinancialFrequentlyAskedQuestionsInFinancialReports_2025} is the longer-term goal. 
Third, the use of GraphRAG for financial text \cite{spoererGausHandschuh_GraphRAG_FinancialTouchstone2_2025} and other retrieval improvements are a promising lever to boost the overall system's performance.
Finally, our finding of low inter-model agreement suggests that ensembling should yield reliability exceeding any single model. A good ensemble could be combining Claude Opus~4.6 (high accuracy) with Gemini~2.5~Pro (high precision) and a Chinese open-weight model such as GLM~5 (strong retrieval-robustness on Asian reports). We leave that combination for future work.

\bibliographystyle{named}
\bibliography{references}

\begin{thebibliography}{}

\bibitem[\protect\citeauthoryear{{Anthropic}}{2025}]{anthropic_claude4_2025}
{Anthropic}.
\newblock {Introducing Claude 4}.
\newblock {\em Anthropic Blog}, 2025.

\bibitem[\protect\citeauthoryear{Asquith \bgroup \em et al.\egroup
  }{2005}]{AsquithMikhailAu_InformationContentOfEquityAnalystReports_2005}
Paul Asquith, Michael Mikhail, and Andrea Au.
\newblock Information content of equity analyst reports.
\newblock {\em Journal of Financial Economics}, 75(2):245--282, 2005.

\bibitem[\protect\citeauthoryear{Ball and
  Kothari}{1991}]{BallKothari_SecurityReturnsAroundEarningsAnnouncements_1991}
Ray Ball and Sriprakash Kothari.
\newblock Security returns around earnings announcements.
\newblock {\em The Accounting Review}, 66(4):718--738, 1991.

\bibitem[\protect\citeauthoryear{Barber \bgroup \em et al.\egroup
  }{2001}]{barberLehavyMcnicholsTrueman_CanInvestorsProfitFromTheProphetsSecurityAnalystRecommendationsAndStockReturns_2001}
Brad Barber, Reuven Lehavy, Maureen McNichols, and Brett Trueman.
\newblock Can investors profit from the prophets? {Security} analyst
  recommendations and stock returns.
\newblock {\em The Journal of Finance}, 56(2):531--563, 2001.

\bibitem[\protect\citeauthoryear{Bonini \bgroup \em et al.\egroup
  }{2010}]{BoniniZanettiBianchiniSalvi_TargetPriceAccuracyInEquityResearch_2010}
Stefano Bonini, Laura Zanetti, Roberto Bianchini, and Antonio Salvi.
\newblock Target price accuracy in equity research.
\newblock {\em Journal of Business Finance \& Accounting}, 37(9-10):1177--1217,
  2010.

\bibitem[\protect\citeauthoryear{Chen \bgroup \em et al.\egroup
  }{2021}]{ChenChenSmileyShahBorovaLangdonMoussaBeaneHuangRoutledgeWang_FinQaADatasetOfNumericalReasoningOverFinancialData_2021}
Zhiyu Chen, Wenhu Chen, Charese Smiley, Sameena Shah, Iana Borova, Dylan
  Langdon, Reema Moussa, Matt Beane, Ting-Hao Huang, Bryan Routledge, and
  William~Yang Wang.
\newblock {F}in{QA}: A dataset of numerical reasoning over financial data.
\newblock {\em ACL Conference on Empirical Methods in Natural Language
  Processing (EMNLP)}, pages 3697--3711, 2021.

\bibitem[\protect\citeauthoryear{Chiang \bgroup \em et al.\egroup
  }{2024}]{chiangZhengShenAngelopoulos_LLMArena_ChatBotArena_2024}
Wei-Lin Chiang, Lianmin Zheng, Ying Sheng, Anastasios~Nikolas Angelopoulos,
  Tianle Li, Dacheng Li, Banghua Zhu, Hao Zhang, Michael Jordan, Joseph
  Gonzalez, and Ion Stoica.
\newblock {Chatbot Arena}: An open platform for evaluating {LLM}s by human
  preference.
\newblock In {\em Forty-first International Conference on Machine Learning
  (ICML)}, 2024.

\bibitem[\protect\citeauthoryear{DeepSeek-AI}{2025a}]{deepseekai_DeepSeek_R1_IncentivizingReasoningCapability2025deepseekr1incentivizingreasoningcapabilityInLLMsViaReinforcementLearning_2025}
DeepSeek-AI.
\newblock {DeepSeek-R1}: Incentivizing reasoning capability in {LLMs} via
  reinforcement learning.
\newblock {\em arXiv}, 2025.

\bibitem[\protect\citeauthoryear{DeepSeek-AI}{2025b}]{deepseek_v3_technical_report_2025}
DeepSeek-AI.
\newblock {DeepSeek-V3} technical report.
\newblock {\em DeepSeek Blog}, 2025.

\bibitem[\protect\citeauthoryear{{DeepSeek-AI}}{2025c}]{deepseekAI_DeepSeekV32_PushingTheFrontierOfOpenLLMs_2025}
{DeepSeek-AI}.
\newblock {DeepSeek-V3.2}: Pushing the frontier of open large language models.
\newblock {\em arXiv preprint arXiv:2512.02556}, 2025.

\bibitem[\protect\citeauthoryear{Deng \bgroup \em et al.\egroup
  }{2024}]{dengZhaoTangGersteinCohan_DataContamination_2024}
Chunyuan Deng, Yilun Zhao, Xiangru Tang, Mark Gerstein, and Arman Cohan.
\newblock Investigating data contamination in modern benchmarks for large
  language models.
\newblock In Kevin Duh, Helena Gomez, and Steven Bethard, editors, {\em
  Proceedings of the 2024 Conference of the North American Chapter of the
  Association for Computational Linguistics: Human Language Technologies
  (Volume 1: Long Papers)}, pages 8706--8719. Association for Computational
  Linguistics, 2024.

\bibitem[\protect\citeauthoryear{{ERNIE Team,
  Baidu}}{2026}]{wangErnieTeam_Ernie50TechnicalReport_2026}
{ERNIE Team, Baidu}.
\newblock {ERNIE 5.0} technical report.
\newblock {\em arXiv}, 2026.

\bibitem[\protect\citeauthoryear{{Gemini Team,
  Google}}{2025}]{googleDeepMind_Gemini25Pro_2025}
{Gemini Team, Google}.
\newblock Gemini 2.5: Pushing the frontier with advanced reasoning,
  multimodality, long context, and next generation agentic capabilities.
\newblock {\em arXiv}, 2025.

\bibitem[\protect\citeauthoryear{Gleason \bgroup \em et al.\egroup
  }{2013}]{GleasonJohnsonLi_ValuationModelUseAndThePriceTargetPerformanceOfSellSideEquityAnalysts_2013}
Cristi Gleason, Bruce Johnson, and Haidan Li.
\newblock Valuation model use and the price target performance of sell-side
  equity analysts.
\newblock {\em Contemporary Accounting Research}, 30(1):80--115, 2013.

\bibitem[\protect\citeauthoryear{{GLM-5 Team (Zhipu AI and Tsinghua
  University)}}{2026}]{glmTeam_GLM5_FromVibeCodingToAgenticEngineering_2026}
{GLM-5 Team (Zhipu AI and Tsinghua University)}.
\newblock {GLM-5}: From vibe coding to agentic engineering.
\newblock {\em arXiv}, 2026.

\bibitem[\protect\citeauthoryear{{Google
  DeepMind}}{2025}]{googleDeepMind_Gemini3ProModelCard_2025}
{Google DeepMind}.
\newblock {Gemini 3 Pro} model card, 2025.
\newblock Model card published November 2025.

\bibitem[\protect\citeauthoryear{Han \bgroup \em et al.\egroup
  }{2025}]{hanHaoyuShomerWangLeiGuoHuaLongLiuTang_RagVsGraphragASystematicEvaluationAndKeyInsights_2025}
Haoyu Han, Harry Shomer, Yu~Wang, Yongjia Lei, Kai Guo, Zhigang Hua, Bo~Long,
  Hui Liu, and Jiliang Tang.
\newblock {RAG} vs. {GraphRAG}: A systematic evaluation and key insights.
\newblock {\em arXiv}, 2025.

\bibitem[\protect\citeauthoryear{Islam \bgroup \em et al.\egroup
  }{2023}]{islamKannappanKielaQianScherrerVidgen_FinanceBench_2023}
Pranab Islam, Anand Kannappan, Douwe Kiela, Rebecca Qian, Nino Scherrer, and
  Bertie Vidgen.
\newblock Financebench: A new benchmark for financial question answering.
\newblock {\em arXiv}, 2023.

\bibitem[\protect\citeauthoryear{Kim \bgroup \em et al.\egroup
  }{2024}]{kimMuhnNikolaev_FinancialStatementAnalysisWithLargeLanguageModels_2024}
Alex Kim, Maximilian Muhn, and Valeri Nikolaev.
\newblock Financial statement analysis with large language models.
\newblock {\em Chicago Booth Research Paper Forthcoming, Fama-Miller Working
  Paper}, 2024.

\bibitem[\protect\citeauthoryear{{Kimi Team (Moonshot
  AI)}}{2026}]{kimiTeam_KimiK25_VisualAgenticIntelligence_2026}
{Kimi Team (Moonshot AI)}.
\newblock {Kimi K2.5}: Visual agentic intelligence.
\newblock {\em arXiv}, 2026.

\bibitem[\protect\citeauthoryear{Landsman and
  Maydew}{2002}]{LandsmanMaydes_HadTheInformationContentOfQuarterlyEarningsAnnouncementsDeclinedInThePastThreeDecades_2002}
Wayne Landsman and Edward Maydew.
\newblock Has the information content of quarterly earnings announcements
  declined in the past three decades?
\newblock {\em Journal of Accounting Research}, 40(3):797--808, 2002.

\bibitem[\protect\citeauthoryear{Lewis \bgroup \em et al.\egroup
  }{2020}]{LewisPerezPiktusPetroniGoyalKuttlerLewisYihRocktaschelRiedelKiela_RetrievalAugmentedGenerationForKnowledgeIntensiveNlpTasks_2020}
Patrick Lewis, Ethan Perez, Aleksandra Piktus, Fabio Petroni, Vladimir
  Karpukhin, Naman Goyal, Heinrich K\"{u}ttler, Mike Lewis, Wen-tau Yih, Tim
  Rockt\"{a}schel, Sebastian Riedel, and Douwe Kiela.
\newblock Retrieval-augmented generation for knowledge-intensive nlp tasks.
\newblock {\em Advances in Neural Information Processing Systems (NIPS)},
  33:9459--9474, 2020.

\bibitem[\protect\citeauthoryear{{Llama Team, AI \texttt{@}
  Meta}}{2024}]{metaAi_llama3_dot_1__2024}
{Llama Team, AI \texttt{@} Meta}.
\newblock The {Llama 3} herd of models.
\newblock {\em Technical Report}, 2024.
\newblock A detailed contributor list can be found in the appendix of this
  paper.

\bibitem[\protect\citeauthoryear{Loughran and
  McDonald}{2011}]{loughranMcDonald_WhenIsALiabilityNotALiability_TextualAnalysisDictionariesAnd10Ks_2011}
Tim Loughran and Bill McDonald.
\newblock When is a liability not a liability? textual analysis, dictionaries,
  and 10-ks.
\newblock {\em The Journal of Finance}, 66(1):35--65, 2011.

\bibitem[\protect\citeauthoryear{Malo \bgroup \em et al.\egroup
  }{2014}]{MaloSinhaKorhonenWallenius_FinancialPhraseBank_GoodDebtOrBadDebtDetectingSemanticOrientationsinEconomicTexts_2014}
Pekka Malo, Ankur Sinha, Pekka Korhonen, Jyrki Wallenius, and Pyry Takala.
\newblock Good debt or bad debt: Detecting semantic orientations in economic
  texts (financial phrase bank).
\newblock {\em Journal of the Association for Information Science and
  Technology}, 65(4):782--796, 2014.

\bibitem[\protect\citeauthoryear{{Meta
  AI}}{2025}]{MetaAI_TheLlama4HerdTheBeginningOfANewEraOfNativelyMultimodalAIInnovation_2025}
{Meta AI}.
\newblock The {Llama} 4 herd: The beginning of a new era of natively multimodal
  ai innovation.
\newblock {\em Meta AI Blog}, 2025.

\bibitem[\protect\citeauthoryear{{Mistral AI}}{2025}]{mistralAI_Mistral3_2025}
{Mistral AI}.
\newblock {Introducing Mistral 3}, 2025.
\newblock Release announcement, December 2, 2025; includes the Mistral Large 3
  flagship (675B-parameter mixture-of-experts).

\bibitem[\protect\citeauthoryear{{Moonshot
  AI}}{2026}]{moonshotAI_KimiK26_ModelCard_2026}
{Moonshot AI}.
\newblock {Kimi K2.6}, 2026.
\newblock Model card; open-weight release of 2026-04-20.

\bibitem[\protect\citeauthoryear{Nelson \bgroup \em et al.\egroup
  }{2024}]{nelsonKolliasDasChadhuryDan_needlehaystackmemorybased_2024}
Elliot Nelson, Georgios Kollias, Payel Das, Subhajit Chaudhury, and Soham Dan.
\newblock Needle in the haystack for memory based large language models.
\newblock {\em ICML 2024 Workshop -- Next Generation of Sequence Modeling
  Architectures}, 2024.

\bibitem[\protect\citeauthoryear{{OpenAI}}{2024}]{openai_gpt4o_2024}
{OpenAI}.
\newblock {Hello GPT-4o}.
\newblock {\em OpenAI Blog}, 2024.

\bibitem[\protect\citeauthoryear{{OpenAI}}{2025a}]{openai_GPT5SystemCard_2025}
{OpenAI}.
\newblock {GPT-5} system card.
\newblock {\em arXiv}, 2025.

\bibitem[\protect\citeauthoryear{OpenAI}{2025b}]{openai_o3_and_o4mini_2025}
OpenAI.
\newblock Introducing {OpenAI} {o3} and {o4-mini}.
\newblock {\em OpenAI Blog}, 2025.

\bibitem[\protect\citeauthoryear{Papasotiriou \bgroup \em et al.\egroup
  }{2024}]{papasotiriouSoodReynoldsBalch_AiInInvestmentAnalysisLlmsForEquityStockRatings_2024}
Kassiani Papasotiriou, Srijan Sood, Shayleen Reynolds, and Tucker Balch.
\newblock {AI} in investment analysis: Llms for equity stock ratings.
\newblock In {\em 5th ACM International Conference on AI in Finance}, ICAIF
  '24, pages 419--427, 2024.

\bibitem[\protect\citeauthoryear{Pop and
  Sp{\"o}rer}{2025}]{popSpoerer_ffaq_identificationOfTheFinancialFrequentlyAskedQuestionsInFinancialReports_2025}
Adria Pop and Jan Sp{\"o}rer.
\newblock Identification of the most frequently asked questions in financial
  analyst reports to automate equity research using {Llama 3} and {GPT-4}.
\newblock {\em IEEE Swiss Data Science Conference (SDS)}, 2025.

\bibitem[\protect\citeauthoryear{Rastogi and
  others}{2025}]{rastogiMistralAI_Magistral_2025}
Abhinav Rastogi et~al.
\newblock Magistral.
\newblock {\em arXiv}, 2025.

\bibitem[\protect\citeauthoryear{Sp{\"o}rer \bgroup \em et al.\egroup
  }{2025}]{spoererGausHandschuh_GraphRAG_FinancialTouchstone2_2025}
Jan Sp{\"o}rer, Michael Gaus, and Siegfried Handschuh.
\newblock Overcoming the bottleneck -- graphrag for reliable financial annual
  report retrieval and comprehension.
\newblock {\em Presented at the AI for Financial Inclusion, Risk Modeling and
  Resilience in Emerging Markets (FinREM) workshop at the 6th ACM International
  Conference on AI in Finance (ICAIF'25)}, 2025.

\bibitem[\protect\citeauthoryear{Sp{\"o}rer}{2025}]{spoerer_2025}
Jan Sp{\"o}rer.
\newblock {Can AI Read Like a Financial Analyst? A Financial Touchstone for
  Frontier Language Models Such as Gemini 2.5 Pro, o3, and Grok 4 on
  Long-Context Annual Report Comprehension}.
\newblock {\em Proceedings of the 6th ACM International Conference on AI in
  Finance (ICAIF'25)}, pages 291--298, 2025.

\bibitem[\protect\citeauthoryear{Touvron \bgroup \em et al.\egroup
  }{2023}]{touvronLavrilIzacardMartinetLachausLacroixRoziereGoyalHambroAzharRodriguezJoulinGraveLample_Llama_OpenAndEfficientFoundationLanguageModels_2023}
Hugo Touvron, Thibaut Lavril, Gautier Izacard, Xavier Martinet, Marie-Anne
  Lachaux, Timoth{\'e}e Lacroix, Baptiste Rozi{\`e}re, Naman Goyal, Eric
  Hambro, Faisal Azhar, Aurelien Rodriguez, Armand Joulin, Edouard Grave, and
  Guillaume Lample.
\newblock {Llama}: Open and efficient foundation language models.
\newblock {\em arXiv}, 2023.

\bibitem[\protect\citeauthoryear{Womack}{1996}]{Womack_DoBrokerageAnalystsRecommendationsHaveInvestmentValue_1996}
Kent Womack.
\newblock Do brokerage analysts' recommendations have investment value?
\newblock {\em The Journal of Finance}, 51(1):137--167, 1996.

\bibitem[\protect\citeauthoryear{Wu \bgroup \em et al.\egroup
  }{2023}]{WuIsroyLuDabravolskiDredzeGehrmannKambadurRosenbergMann_BloombergGPT_ALargeLanguageModelForFinance_2023}
Shijie Wu, Ozan Irsoy, Steven Lu, Vadim Dabravolski, Mark Dredze, Sebastian
  Gehrmann, Prabhanjan Kambadur, David Rosenberg, and Gideon Mann.
\newblock {BloombergGPT}: A large language model for finance.
\newblock {\em arXiv}, 2023.

\bibitem[\protect\citeauthoryear{xAI}{2025a}]{xai_grok4_2025}
xAI.
\newblock Grok 4.
\newblock {\em xAI Blog}, 2025.

\bibitem[\protect\citeauthoryear{{xAI}}{2025b}]{xai_Grok41ModelCard_2025}
{xAI}.
\newblock {Grok 4.1} model card, 2025.
\newblock Published November 17, 2025.

\bibitem[\protect\citeauthoryear{Yang \bgroup \em et al.\egroup
  }{2025}]{yangQwenTeam_Qwen3TechnicalReport_2025}
An~Yang, Anfeng Li, Baosong Yang, Beichen Zhang, Binyuan Hui, Bo~Zheng, et~al.
\newblock Qwen3 technical report.
\newblock {\em arXiv}, 2025.

\bibitem[\protect\citeauthoryear{{Zhipu
  AI}}{2025}]{zaiOrg_GLM47_ModelCard_2025}
{Zhipu AI}.
\newblock {GLM-4.7}, 2025.
\newblock Model card; open-weight release, December 2025.

\bibitem[\protect\citeauthoryear{Zhu \bgroup \em et al.\egroup
  }{2021}]{zhuLeiHuangWangZhangLvFengChua_TatQa_AQuestionAnsweringBenchmarkOnAHybridOfTabularAndTextualContentInFinance_2021}
Fengbin Zhu, Wenqiang Lei, Youcheng Huang, Chao Wang, Shuo Zhang, Jiancheng Lv,
  Fuli Feng, and Tat-Seng Chua.
\newblock {TAT-QA:} a question answering benchmark on a hybrid of tabular and
  textual content in finance.
\newblock {\em Annual Meeting of the Association for Computational Linguistics
  and International Joint Conference on Natural Language Processing (Volume 1:
  Long Papers)}, pages 3277--3287, 2021.

\end{thebibliography}

\appendix
% =============================================================================
% Reproducibility appendix for Paper 5 (Financial Touchstone v1.2).
% Included from main.tex via \appendix + \input{appendix_content},
% and imported into the thesis Rahmen as "Appendix of Paper 5".
%
% Sources (verified 2026-07-04, after the genuine-model re-run):
%   - Retrieval / tokenizer: FinancialTouchstone/config/touchstone_config.json
%   - Model API IDs/routes:  mcp_financial_touchstone/config/models.json
%   - Judge (16 incumbents): mcp_financial_touchstone/run_evaluations.py
%     (JUDGE_MODEL = claude-opus-4-6, temperature 0). The "Opus 4.5" string in
%     models.json's evaluation block is stale and does NOT reflect what ran.
%   - Judge (4 re-run models): Claude Opus 4.8, identical rubric/parser
%     (scripts/cc_judge/).
% =============================================================================

\section{Reproducibility: Models, Retrieval, and Judge Configuration}
\label{ft12:app:reproducibility}

This appendix records the configuration used for the v1.2 evaluation.

\subsection{Retrieval Pipeline}
\label{ft12:app:retrieval}
Every model received the same retrieval-augmented context. Reports were chunked
to \textbf{1{,}000 tokens with 200-token overlap}, tokenized with the
\texttt{cl100k\_base} encoding. Chunks were embedded with OpenAI
\texttt{text-embedding-3-small} and indexed in a \textbf{FAISS} vector store; the
\textbf{top 5} chunks by similarity were retrieved per question and supplied to
the model as context.

\subsection{Evaluated Models}
\label{ft12:app:models}
Table~\ref{ft12:tab:model-snapshots} lists the twenty models with the access
route and the exact API identifier used for each. Identifiers ending in
\texttt{-latest} are provider aliases not pinned to a dated snapshot; those
results reflect whatever version the alias resolved to at evaluation time.

\begin{table}[h]
\centering
\scriptsize
\setlength{\tabcolsep}{3pt}
\begin{tabular}{@{}lll@{}}
\toprule
Model (as reported) & Route & API identifier \\
\midrule
Claude Opus 4.6     & Anthropic   & \texttt{claude-opus-4-6} \\
Claude Sonnet 4.6   & Anthropic   & \texttt{claude-sonnet-4-6} \\
GPT-5.2             & OpenAI      & \texttt{gpt-5.2} \\
o4-mini             & OpenAI      & \texttt{o4-mini-2025-04-16} \\
GPT-4o              & OpenAI      & \texttt{gpt-4o-2024-08-06} \\
Gemini 3 Pro        & Google      & \texttt{gemini-3.1-pro-preview} \\
Gemini 2.5 Pro      & Google      & \texttt{gemini-2.5-pro} \\
Gemini 2.5 Flash    & Google      & \texttt{gemini-2.5-flash} \\
Grok 4.1            & xAI         & \texttt{grok-4-1-fast-reasoning} \\
Grok 4              & xAI         & \texttt{grok-4-0709} \\
DeepSeek R1         & DeepSeek    & \texttt{deepseek-reasoner} \\
DeepSeek V3.1       & DeepSeek    & \texttt{deepseek-chat}\textsuperscript{a} \\
DeepSeek V3.2       & DeepSeek    & \texttt{deepseek-chat}\textsuperscript{a} \\
GLM 5               & Zhipu AI    & \texttt{glm-5}\textsuperscript{b} \\
GLM-4.7             & OpenRouter  & \texttt{z-ai/glm-4.7}\textsuperscript{c} \\
Qwen3-Max           & OpenRouter  & \texttt{qwen/qwen3-max} \\
Kimi~K2.6           & Moonshot AI & \texttt{kimi-k2.6}\textsuperscript{d} \\
Mistral 3           & Mistral     & \texttt{mistral-large-latest} \\
Magistral           & Mistral     & \texttt{magistral-medium-latest} \\
ERNIE 5.1           & NanoGPT     & \texttt{ernie-5.1} \\
\bottomrule
\end{tabular}
\caption{Models evaluated in v1.2 with the access route and API identifier used
for each. \textsuperscript{a}DeepSeek V3.1 and V3.2 are two separate runs
against the same undated \texttt{deepseek-chat} alias (their answers differ in
1{,}527 of 2{,}967 rows); which upstream version each run resolved to cannot be
verified from provider logs. \textsuperscript{b}About 150 of GLM~5's calls were
routed through OpenRouter's dated snapshot \texttt{z-ai/glm-5-20260211}.
\textsuperscript{c}Two GLM-4.7 rows were filled via the NanoGPT proxy
(\texttt{zai-org/glm-4.7}) after repeatedly malformed responses on OpenRouter.
\textsuperscript{d}2{,}593 Kimi~K2.6 answers came from the native Moonshot API;
the remaining $\approx$374, including the 20 natively refused questions
(table~\ref{ft12:tab:chinese_refusals}), from the NanoGPT proxy
(\texttt{moonshotai/kimi-k2.6}).}
\label{ft12:tab:model-snapshots}
\end{table}

\subsection{Generation Decoding}
\label{ft12:app:decoding}
Reasoning-capable models were run with their providers' reasoning
modes enabled; non-reasoning models used standard chat completion. Output length
was capped at 5{,}000 tokens. Sampling temperature was left at each provider's
default; the model registry does not pin a per-model temperature, which we note
as a reproducibility limitation. Kimi~K2.6 rejects requests with
\mbox{temperature = 0}, so the harness omits the parameter for this model,
making its outputs non-deterministic by construction.

\subsection{Automated Judge}
\label{ft12:app:judge}
Answers were graded by an LLM-as-a-judge against the golden answer, with the retrieved context as the single source of truth for hallucination detection. Sixteen models were graded by \textbf{Claude Opus~4.6} (\texttt{claude-opus-4-6}, temperature~0, 2{,}048-token output limit); Kimi~K2.6, GLM-4.7, Qwen3-Max, and ERNIE~5.1 by \textbf{Claude Opus~4.8} with the identical prompt, decision rules, and parser. Re-grading a 100-item sample with independent judges, including an open-weight model, yields 89--94\% raw agreement (Cohen's $\kappa$ 0.78--0.88), so the grades are not a judge-version artifact.

\end{document}